\documentclass{article}

\usepackage{amsmath,amssymb}
\usepackage[mathscr]{euscript} 
\usepackage{amsthm} 
\usepackage{tikz} 
\usepackage{graphicx} 
\newcommand*{\Scale}[2][4]{\scalebox{#1}{\ensuremath{#2}}} 
\usepackage{framed} 
\usepackage{float} 
\usepackage{multirow}
\usepackage{color}
\usepackage{subfigure}
\usepackage{longtable}
\usepackage{cite}
\usepackage[hidelinks, breaklinks=true]{hyperref} 
\usepackage{xcolor}  \definecolor{shadecolor}{rgb}{.95,.95,.95}  
\usepackage[font=footnotesize]{caption}
\usepackage{nicefrac}
\usepackage{dsfont} 
\usepackage[export]{adjustbox} 
\usepackage{cancel} 
\usepackage{enumitem}	

\usepackage[ruled,vlined]{algorithm2e}

\theoremstyle{definition}

\tikzstyle{every edge}=  [draw]
\tikzstyle{vertex} = [draw,circle,minimum size=1pt]
\tikzstyle{label} = [minimum size=.1pt,font=\scriptsize]
\tikzstyle{title} = [minimum size=.25cm,font=\small]

\newcommand{\bs}[1]{\boldsymbol{#1}}
\newcommand{\bhat}[1]{\boldsymbol{\hat{#1}}}

\newcommand{\gray}{\textcolor{gray}}

\def \R{\mathbb{R}}

\def \1{{\mathds{1}}}
\def \T{\mathsf{T}}

\def \spn{{\rm span}}

\def \<{\langle}
\def \>{\rangle}

\DeclareMathOperator*{\argmin}{arg\,min}

\def \Fr{{\hyperref[FrDef]{{\rm F}}}}

\def \K{{\hyperref[KDef]{{\rm K}}}}
\def \n{{\hyperref[nDef]{{\rm n}}}}
\def \d{{\hyperref[dDef]{{\rm d}}}}
\def \r{{\hyperref[rDef]{{\rm r}}}}
\def \xi{{\hyperref[xiDef]{{\rm x}}}}
\def \p{{\hyperref[pDef]{{\rm p}}}}
\def \lambdaa{{\hyperref[lambdaaDef]{\lambda}}}
\def \sigmaa{{\hyperref[sigmaaDef]{\sigma}}}

\def \x{{\hyperref[xDef]{\bs{{\rm x}}}}}
\def \y{{\hyperref[yDef]{\bs{{\rm y}}}}}
\def \bmu{{\hyperref[bmuDef]{\bs{\mu}}}}
\def \btheta{{\hyperref[bthetaDef]{\bs{\theta}}}}

\def \I{{\hyperref[IDef]{\bs{{\rm I}}}}}

\def \X{{\hyperref[XDef]{\bs{{\rm X}}}}}
\def \U{{\hyperref[UDef]{\bs{{\rm U}}}}}
\def \P{{\hyperref[PDef]{\bs{{\rm P}}}}}
\def \bnabla{{\hyperref[bnablaDef]{\bs{\nabla}}}}
\def \bTheta{{\hyperref[bThetaDef]{\bs{\Theta}}}}
\def \S{{\hyperref[SDef]{\bs{{\rm S}}}}}
\def \W{{\hyperref[WDef]{\bs{{\rm W}}}}}

\def \i{{\hyperref[iDef]{{\rm i}}}}
\def \j{{\hyperref[jDef]{{\rm j}}}}
\def \k{{\hyperref[kDef]{{\rm k}}}}

\def \O{{\hyperref[ODef]{\Omega}}}
\def \o{{\hyperref[oDef]{\omega}}}

\def \sU{{\hyperref[sUDef]{\mathbb{U}}}}

\def \FSC{{\hyperref[FSCDef]{{\sc Fsc}}}}
\def \SC{{\hyperref[SCDef]{SC}}}
\def \SSC{{\hyperref[SSCDef]{SSC}}}
\def \UoS{{\hyperref[UoSDef]{UoS}}}
\def \LRMC{{\hyperref[LRMCDef]{LRMC}}}
\def \PCA{{\hyperref[PCADef]{PCA}}}

\newcommand*{\titleGP}{\begingroup 
\centering 
\vspace*{\baselineskip} 

\rule{\textwidth}{1.6pt}\vspace*{-\baselineskip}\vspace*{2pt} 
\rule{\textwidth}{0.4pt}\\[.5\baselineskip] 

{\LARGE Fusion Subspace Clustering: \\ [0.5\baselineskip] Full \& Incomplete Data} \\ [0.4\baselineskip] 

\rule{\textwidth}{0.4pt}\vspace*{-\baselineskip}\vspace{3.2pt} 
\rule{\textwidth}{1.6pt}\\[\baselineskip] 

{\scshape 
Daniel L. Pimentel-Alarc\'on, Usman Mahmood}

{\itshape Georgia State University\par}

{\texttt{pimentel@gsu.edu}, \texttt{umahmood1@student.gsu.edu}\par}
\endgroup}

\begin{document}
\titleGP

\begin{abstract}
\sloppypar Modern inference and learning often hinge on identifying low-dimensional structures that approximate large scale data. Subspace clustering achieves this through a union of linear subspaces. However, in contemporary applications data is increasingly often incomplete, rendering standard (full-data) methods inapplicable. On the other hand, existing incomplete-data methods present major drawbacks, like lifting an already high-dimensional problem, or requiring a super polynomial number of samples. Motivated by this, we introduce a new subspace clustering algorithm inspired by fusion penalties. The main idea is to permanently assign each datum to a subspace of its own, and minimize the distance between the subspaces of all data, so that subspaces of the same cluster get {\em fused} together. Our approach is entirely new to both, full and missing data, and unlike other methods, it directly allows noise, it requires no liftings, it allows low, high, and even full-rank data, it approaches optimal (information-theoretic) sampling rates, and it does not rely on other methods such as low-rank matrix completion to handle missing data. Furthermore, our extensive experiments on both real and synthetic data show that our approach performs comparably to the state-of-the-art with complete data, and dramatically better if data is missing.
\end{abstract}

\section{Introduction}
Inferring low-dimensional structures that explain high-dimensional data has become a cornerstone of discovery in virtually all fields of science. \phantomsection\label{PCADef}Principal component analysis (\PCA), which identifies the low-dimensional linear subspace that best explains a dataset, is arguably the most prominent technique for this purpose. However, in many applications \---- computer vision, image processing, bioinformatics, linguistics, networks analysis, and more \cite{kanade, charRecog, kanatani, lambertian, recommender, scc, eriksson, guessWho, ssc, network} \---- data is often composed of a mixture of several classes, each of which can be explained with a different subspace. Clustering and inferring subspaces that explain data is an important unsupervised learning problem that has received tremendous attention in recent years, producing theory and algorithms to handle outliers, noisy measurements, privacy concerns, and data constraints, among other difficulties \cite{scVidal, liu1, liu2, mahdi, qu, peng, wang, aarti, hu, scalableSC, L0sparse, dataDependent}.

However, one major challenge in contemporary problems is that data is often incomplete. For example, in image inpainting, the values of some pixels are missing due to faulty sensors and image contamination \cite{inpainting}; in computer vision features are often missing due to occlusions and tracking algorithms malfunctions \cite{occlusions}; in recommender systems each user only rates a limited number of items \cite{collaborativeRanking}; in a network, most nodes communicate in subsets, producing only a handful of all the possible measurements \cite{eriksson}.


{\bf Prior Work.} There are numerous approaches to subspace clustering with missing data. For example, \cite{ewzf} suggests using the standard (full-data) state-of-the-art algorithm \phantomsection\label{SSCDef}sparse subspace clustering (\SSC) \cite{ssc} after filling in the missing entries with a sensible value (e.g., zeros or means) or with \phantomsection\label{LRMCDef}low-rank matrix completion (\LRMC) \cite{candes-recht}. However, the number and dimensions of the subspaces are often large enough that \LRMC\ methods are not applicable, and data filled with zeros or means no longer lie in a \phantomsection\label{UoSDef}union of subspaces (\UoS), thus guaranteeing failure even with a modest amount of missing data \cite{elhamifar}. On the other hand, \cite{hrmc} gives an algorithm that uses partial neighborhoods and provably works, but requires a super-polynomial amount of samples, which is unusual in most applications. Alternating methods include adaptations of $k$-subspaces to handle missing data \cite{kGROUSE}, an expectation-maximization (EM) algorithm that models \UoS\ as a gaussian mixture \cite{ssp14}, and a group-lasso formulation \cite{gssc}. However, these alternating algorithms depend heavily on initialization, and can only be guaranteed to converge to a local minimum. More recently, \cite{elhamifar,greg,ladmc} use lifting techniques, yet this requires (at the very least) squaring the dimension of an already high-dimensional problem, which severely limits their applicability.

{\bf Paper Contributions.} In this paper we propose an entirely different approach to \phantomsection\label{SCDef}subspace clustering (\SC), inspired by fusion penalties \cite{fusionVariable, fusedLasso, groupPursuit, clusterpath, sumofnorms, fusion}. The main idea is to permanently assign each datum to a subspace of its own, and then {\em fuse} together nearby subspaces by minimizing (i) the distance between each datum and its subspace (thus guaranteeing that each datum is explained by its subspace), and (ii) the distance between the subspaces from all data, so that subspaces from points that belong together get fused into one. Our algorithm, which we call \phantomsection\label{FSCDef}{\em fusion subspace clustering} (\FSC) is new to both, the full and incomplete data settings, and has the next advantages over the state-of-the-art:
\begin{itemize}[leftmargin=0.4cm]
\item[--]
\FSC\ does not require a super-polynomial number of samples, as \cite{hrmc}. In fact, our experiments show that \FSC\ succeeds with only a few more samples than the strictly necessary \cite{infoTheoretic}.
\item[--]
There is a natural way to extend \FSC\ to handle missing data, unlike other algorithms, including the state-of-the-art for full data, \SSC. Similarly, \FSC\ carries unchanged to noisy settings, unlike other results such as \cite{infoTheoretic}.
\item[--]
\LRMC\ methods adapted to multiple subspaces, as well as \SC\ methods adapted to missing data often rely on \SSC\ and \LRMC\ \cite{ewzf,gssc,elhamifar,greg,ladmc}. In contrast, \FSC\ is a novel algorithm, rather than an adaptation, and it does not rely on \SSC\ nor any other \SC\ algorithm. In addition to clustering, \FSC\ provides a subspace estimation method, and a data completion method that do not require \LRMC.
\item[--]
Two-stage methods like \cite{ewzf}, which first complete using \LRMC, and then cluster using \SSC\ (or vice versa), only work if data lie in a \UoS\ but remains low-rank, allowing only a few number of subspaces of very small dimensions.  \FSC\ works even in the more general setting where data lie in a \UoS\ and is high-rank, or even full-rank, thus allowing more subspaces and of larger dimensions.
\item[--]
Unlike alternating algorithms such as \cite{ssp14,gssc}, \FSC\ does not require knowing the number of subspaces nor their dimensions. Similar to hierarchical clustering, \FSC\ can produce a progressive clustering that gradually fuses subspaces together. Combined with a simple goodness of fit test, like the Akaike information criterion (AIC) \cite{akaike}, this provides a model selection criteria to determine the number of subspaces and their dimensions.
\item[--]
Recent approaches such as \cite{elhamifar,greg,ladmc} exploit the algebraic structure of a \UoS. In fact, each \UoS\ describes an algebraic (non-linear) variety that can be used to cluster and complete. However, expressing the polynomials of these varieties requires lifting (or tensorizing) the data, which turns an already high-dimensional problem into an extremely high-dimensional one, thus limiting their applicability. For example, processing a small image of $64\times64$ pixels requires a vector space of dimension $\d=64{}^2=4,096$.  Even the smallest lifting results in a space of dimension ${\d+1 \choose 2}=8,390,656$. In contrast, \FSC\ requires no data liftings.
\item[--]
Our experiments on real and synthetic data show that with full data, \FSC\ performs comparably to the state-of-the-art, and dramatically better if data is missing.
\end{itemize}

\section{Problem Statement}
\label{problemSec}
Let \phantomsection\label{XDef}\phantomsection\label{dDef}\phantomsection\label{nDef}$\X \in \R{}^{\d \times \n}$ be a data matrix whose columns lie {\em approximately} in the union of \phantomsection\label{KDef}$\K$ low-dimensional subspaces of $\R{}^\d$ (i.e., we allow noise).  Assume that we do not know a priori the subspaces, nor how many there are, nor their dimensions, nor which column belongs to which subspace.  Let $\X{}^\O$ denote the incomplete version of $\X$, observed only in the entries of \phantomsection\label{ODef}$\O \subset \{1,\dots,\d\} \times \{1,\dots,\n\}$.  Given $\X{}^\O$, our goals are to cluster the columns of $\X{}^\O$ according to the underlying subspaces, infer such subspaces, and complete $\X{}^\O$.

{\bf Notations.} Throughout the paper, \phantomsection\label{xDef}$\x_\i \in \R{}^\d$ denotes the $\i{}^{\rm th}$ column of $\X$, \phantomsection\label{sUDef}$\sU_\i \subset \R{}^\d$ denotes the subspace assigned to $\x_\i$, and \phantomsection\label{UDef}$\U_\i \in \R{}^{\d \times \r}$ is a basis of $\sU_\i$; here \phantomsection\label{iDef}$\i=1,\dots,\n$, and \phantomsection\label{rDef}$\r$ is an upper bound on the dimension of the subspaces. Given $\i$, we use the superscript \phantomsection\label{oDef}$\o$ to indicate the restriction of a subspace, matrix or vector to the observed entries in $\x_\i$. For example, if $\x_\i$ is observed on $\ell$ entries, then $\x{}_{\i}^\o \in \R{}^{\ell}$ and $\U{}_{\i}^\o \in \R{}^{\ell \times \r}$ denote the restrictions of $\x_\i$ and $\U_\i$ to the observed entries in $\x_\i$. Similarly, we denote the projection operators onto $\sU_\i$ and $\sU{}_{\i}^\o$ (the subspace assigned to $\x_\i$, and its restriction to the observed entries in $\x_\i$) as\phantomsection\label{PDef}
\begin{align}
\label{projectionEq}
\P_\i:=\U_\i (\U_\i^\T \U_\i)^{-1}\U_\i^\T \in \R^{\d \times \d}
\hspace{.75cm} \text{and} \hspace{.75cm}
\P_{\i}^\o:=\U_{\i}^\o (\U_{\i}^{\o\T} \U_{\i}^\o)^{-1}\U_{\i}^{\o\T} \in \R^{\ell \times \ell}.
\end{align}
Finally, \phantomsection\label{FrDef}$\|\bs{\cdot}\|{}_\Fr^2$ denotes the squared Frobenius norm, given by the sum of squared entries in a matrix.

\section{Review of Fusion Penalties}
\label{reviewSec}
Fusion penalties can be viewed as a convex relaxation of hierarchical clustering \cite{fusionVariable, fusedLasso, groupPursuit, clusterpath, sumofnorms, fusion}. Given $\n$ points \phantomsection\label{yDef}$\y_1,\dots,\y_\n \in \R{}^\d$ (not necessarily in a \UoS), hierarchical clustering is the greedy algorithm that recursively joins nearby points until only $\K$ clusters remain, aiming to find:\phantomsection\label{bmuDef}
\begin{align}
\label{hierarchicalEq}
\argmin_{\bmu_1,\dots,\bmu_\n} \ \ \sum_{\i=1}^\n \| \y_\i-\bmu_\i \|_2^2
\hspace{.5cm} \text{subject to} \hspace{.5cm}
\frac{1}{2} \sum_{\i=1}^\n \sum_{\j=1}^\n \1_{\{\bmu_\i \neq \bmu_\j\}} \leq \K,
\end{align}
where \phantomsection\label{1Def}$\1$ denotes the indicator function. Notice that if $\K \geq \n(\n+1)/2$ (the number of distinct pairs of points), then the problem is unconstrained, and the trivial solution is $\y_\i=\bmu_\i$ for every $\i$. If $\K=\n(\n+1)/2-1$, then \eqref{hierarchicalEq} forces two centers to {\em fuse}, which equates to the first step in hierarchical clustering. More generally, if $\K=\n(\n+1)/2-\ell$, then \eqref{hierarchicalEq} forces $\ell-1$ centers to fuse. Since \eqref{hierarchicalEq} is a difficult combinatorial problem, \cite{clusterpath} proposed the following convex relaxation:
\begin{align}
\label{clusterpathEq}
\argmin_{\bmu_1,\dots,\bmu_\n} \ \
\sum_{\i=1}^\n \| \y_\i-\bmu_\i \|_2^2
\hspace{.5cm} \text{subject to} \hspace{.5cm}
\frac{1}{2} \sum_{\i=1}^\n \sum_{\j=1}^\n \| \bmu_\i - \bmu_\j \|_2^2 \leq \K'.
\end{align}
Intuitively, \eqref{clusterpathEq} aims to minimize the euclidian distance between each point $\y_\i$ and its assigned center $\bmu_\i$, while at the same time guaranteeing that the centers from different points are not too far from one an other (which is regulated by $\K'$). The next section uses these insights to address subspace clustering.  The crucial difference is that now {\em centers} will be subspaces rather than points, and so we can no longer use the euclidian distances in \eqref{clusterpathEq}, as two points in the same subspace can be arbitrarily apart with respect to the euclidian distance.

\section{Fusion Subspace Clustering}
\label{fscSec}
We now present our novel algorithm: Fusion Subspace Clustering (\FSC). The main idea is to permanently assign each column $\x_\i$ to a subspace of its own that is close to $\x_\i$, and close to subspaces assigned to other columns, so that the subspaces that belong together get {\em fused}. To do this, we will minimize (i) the residual of $\x_\i$ when projected onto $\sU_\i$, to measure the distance between $\x_\i$ and its assigned subspace $\sU_\i$, and (ii) the difference between the projector operators of $\sU_\i$ and $\sU_\j$, to measure the distance (over the Grassmannian) between the subspaces assigned to columns $\i$ and \phantomsection\label{jDef}$\j$, i.e.,
\begin{framed}
\begin{align}
\label{fscEq}
\argmin_{\U_1,\dots,\U_\n} \ \ \sum_{\i=1}^\n \|\x_\i-\P_{\i} \x_{\i} \|_2^2
\ + \
\frac{\lambdaa}{2} \sum_{\i=1}^\n \sum_{\j=1}^\n \| \P_\i-\P_\j \|_\Fr^2,
\end{align}
\end{framed}
where the dependency on each $\U_\i$ is hidden in $\P_\i$ (see \eqref{projectionEq}), and \phantomsection\label{lambdaaDef}$\lambdaa \geq 0$ is a proxy of $\K$ that regulates how clusters fuse together. The larger $\lambdaa$, the more we penalize subspaces being apart, which results in subspaces getting closer (fused); more details in Section \ref{modelSelectionSec}. Recall that $\U_\i \in \R{}^{\d \times \r}$, where $\r$ is an upper bound on the subspaces dimension. Section \ref{modelSelectionSec} also includes more details about estimating $\r$.

In order to solve \eqref{fscEq}, in our experiments we use a random initialization and standard gradient descent, where the gradient of $\U_\i$ is given by:\phantomsection\label{bnablaDef}
\begin{align}
\label{gradientEq}
\bnabla_\i \ &= \ -2 \x_\i\x_\i^\T \U_\i (\U_\i^\T \U_\i)^{-1} 
\ + \ (\U_\i\U_\i^\T)^2\x_\i\x_\i^\T\U_\i(\U_\i^\T\U_\i)^{-1} \nonumber \\
\ &+ \ \U_\i (\U_\i^\T\U_\i)^{-1} \U_\i^\T\x_\i\x_\i^\T\U_\i \U_\i^\T\U_\i \nonumber \\
\ &+ \ 8 \sum_{\j \neq \i} (\U_\i(\U_\i^\T\U_\i)^{-1}\U_\i^\T-\I)\U_\j(\U_\j^\T\U_\j)^{-1}\U_\j^\T \U_\i(\U_\i^\T\U_\i)^{-1}.
\end{align}
However, we can also solve \eqref{fscEq} using more sophisticated techniques, like an alternating direction method of multipliers (ADMM) formulation \cite{admm1,admm2} and smarter initializations, for example setting the initial subspaces to be orthogonal or with small principal angles between them.

The solution to \eqref{fscEq} will be a sequence of subspace bases $\U_1,\dots,\U_\n$, one for each column in $\X$. Due to the second term in \eqref{fscEq}, we expect subspace $\sU_\i:=\spn\{\U_\i\}$ to be close to $\sU_\j:=\spn\{\U_\j\}$ if columns $\x_\i$ and $\x_\j$ belong together, and far otherwise. It remains to group together subspaces that are close, or equivalently, assign a label to each subspace $\sU_\i$. To this end, use spectral clustering, which shows remarkable performance in many modern problems, and is widely used as the final step in many subspace clustering algorithms, including the state-of-the-art \SSC.
Spectral clustering receives a similarity matrix \phantomsection\label{SDef}$\S \in \R{}^{\n \times \n}$ between $\n$ points, and runs a standard clustering method (like $k$-means) on the relevant eigenvectors of the Laplacian matrix of $\S$ \cite{}. We can build a similarity matrix $\S$ between subspaces $\sU_1,\dots,\sU_\n$ whose $(\i,\j){}^{\rm th}$ entry is equal to $1/\| \P_\i-\P_\j \|{}_\Fr^2$. At this point we can run spectral clustering on $\S$ to assign a label \phantomsection\label{kDef}$\k_\i \in \{1,\dots,\K\}$ to each subspace $\sU_\i$, or equivalently, to each column $\x_\i$, thus providing a clustering of $\X$, as desired. The entire process of \FSC\ is summarized in Algorithm \ref{fscAlg}.

\section{A Natural Generalization to Missing Data}
\label{missingSec}
Section \ref{fscSec} introduces \FSC\ in its full-data setting. If data is missing, the only difference is that each subspace only needs to explain the observed entries of its assigned column, resulting in the following:
\begin{framed}
\begin{align}
\label{ifscEq}
\argmin_{\U_1,\dots,\U_\n} \ \ \sum_{\i=1}^\n \|\x_\i^{\o}-\P^\o_{\i} \x^\o_{\i} \|_2^2
\ + \
\frac{\lambdaa}{2} \sum_{\i=1}^\n \sum_{\j=1}^\n \| \P_\i-\P_\j \|_\Fr^2.
\end{align}
\end{framed}
In words, \eqref{ifscEq} simply ignores all the unobserved entries in the first term of \eqref{fscEq}.
Notice that if columns $\x{}_\i^\o$ and $\x{}_\j^\o$ belong together, we still want the whole subspaces $\sU_\i$ and $\sU_\j$ to be close, so the second term in \eqref{fscEq} remains unchanged. As consequence, this time the gradient of $\U_\i$ is equal to
\begin{align*}
\bnabla_\i \ = \ \bnabla_\i' \ + \ 8 \sum_{\j \neq \i} (\U_\i(\U_\i^\T\U_\i)^{-1}\U_\i^\T-\I)\U_\j(\U_\j^\T\U_\j)^{-1}\U_\j^\T \U_\i(\U_\i^\T\U_\i)^{-1},
\end{align*}
where $\bnabla_\i'$ is equal to $\bs{0}$ for the rows not in $\o$, and equal to
\begin{align*}
\bnabla_\i^\o \ = \ &-2 \x_\i^\o \x_\i^{\o\T} \U_\i^\o (\U_\i^{\o\T} \U_\i^\o)^{-1} 
\ + \ (\U_\i^\o\U_\i^{\o\T})^2\x_\i^\o\x_\i^{\o\T}\U_\i^\o(\U_\i^{\o\T}\U_\i^\o)^{-1} \\
&+ \ \U_\i^\o (\U_\i^{\o\T}\U_\i^\o)^{-1} \U_\i^{\o\T}\x_\i^\o\x_\i^{\o\T}\U_\i^\o \U_\i^{\o\T}\U_\i^\o.
\end{align*}
for the rows in $\o$. Intuitively, the incomplete-data gradient is equal to \eqref{gradientEq}, only ignoring the first line for the rows not in $\o$.  Using this new gradient, we can solve \eqref{ifscEq} same as \eqref{fscEq}, using random or orthogonal initializations and gradient descent, or more sophisticated methods, like ADMM \cite{admm1,admm2}. We point out that even though $\X{}^\O$ has missing data, the solution to \eqref{ifscEq} will be a sequence of (full) subspace bases $\U_1,\dots,\U_\n$, same as in the full-data case. Hence, as the final step we can assign labels using spectral clustering, same as before, thus clustering $\X{}^\O$.

\begin{algorithm}[b]
\caption{Fusion Subspace Clustering.}
\label{fscAlg}

\begin{minipage}{0.45\textwidth}
\begin{center}
FULL DATA:
\end{center}
\Scale[.7]{\gray{1}}. \textbf{Input:} Columns $\x_1,\dots,\x_\n$.

\Scale[.7]{\gray{2}}. Solve \eqref{fscEq} to obtain $\U_1,\dots,\U_\n$.
\end{minipage}
\hfill
\begin{minipage}{0.05\textwidth}
\begin{tabular}{||p{\textwidth}}
\\ \\ \\ \\
\end{tabular}
\end{minipage}
\hfill
\begin{minipage}{0.4\textwidth}
\begin{center}
INCOMPLETE DATA:
\end{center}
\hspace{-.5cm}
\Scale[.7]{\gray{1}}. \textbf{Input:} Columns $\x^\o_1,\dots,\x_\n^\o$.

\hspace{-.5cm}
\Scale[.7]{\gray{2}}. Solve \eqref{ifscEq} to obtain $\U_1,\dots,\U_\n$.
\end{minipage}

\begin{minipage}{0.1\textwidth}
\hspace{.05cm}
\end{minipage}
\begin{minipage}{0.7\textwidth}
\Scale[.7]{\gray{3}}. Compute $\S$, the similarity matrix of $\U_1,\dots,\U_\n$.

\Scale[.7]{\gray{4}}. Spectral cluster $\S$ to obtain labels $\k_1,\k_2,\dots,\k_\n$.

\Scale[.7]{\gray{5}}. \textbf{Output:} The $\i^{\rm th}$ column corresponds to cluster $\k_\i$.
\end{minipage}
\hfill
\end{algorithm}

\subsection{Estimating Subspaces and Completing Data}
\label{completionSec}
Recall that our goals are to: (i) cluster the columns of $\X{}^\O$, (ii) infer the underlying subspaces, and (iii) complete $\X{}^\O$. So far we have only achieved (i). However, that is the difficult step. In fact, once $\X{}^\O$ is clustered, there are several straightforward ways to achieve (ii) and (iii).

Common approaches concatenate all the columns of $\X{}^\O$ that correspond to the same cluster into a single matrix $\X{}^\O_{\k}$, and complete it into a matrix $\bhat{\X}_\k$ using \LRMC\ (because its columns now lie in a single subspace), thus achieving (iii). To accomplish (ii) one can compute the leading singular vectors of $\bhat{\X}_\k$ to produce an subspace basis estimate $\bhat{\U}_\k$. We can do this as well. However \FSC\ offers natural estimation and completion methods that do not rely on \LRMC, which may fail if the subspaces are coherent (somewhat aligned with the canonical axes) or samples are not uniformly spread \cite{candes-recht}.

Our \LRMC-free approach is as follows: since the bases $\U_1,\dots,\U_\n$ produced by \eqref{ifscEq} have no missing data, we can normalize and concatenate all the bases that correspond to the $\k{}^{\rm th}$ cluster into a single matrix \phantomsection\label{WDef}$\W_\k$, and compute its leading singular vectors to produce an {\em average} estimate $\bhat{\U}_\k$, thus achieving (ii). Next we can estimate the coefficient of each incomplete column $\x{}_\i^\o$ with respect to its corresponding subspace basis $\bhat{\U}_{\k_\i}$ as \phantomsection\label{bthetaDef}$\bhat{\btheta}_\i := (\bhat{\U}{}_{\k_\i}^{\o\T} \bhat{\U}{}_{\k_\i}^\o)^{-1}\bhat{\U}{}_{\k_\i}^{\o\T}\x{}_\i^\o$. Since the coefficient of $\x{}_\i^\o$ is the same as the coefficient of $\x_\i$ we can complete $\x{}_\i^\o$ as $\bhat{\x}_\i=\bhat{\U}_{\k_\i}\bhat{\btheta}_\i$, thus achieving (iii).

\section{Model Selection}
\label{modelSelectionSec}
\begin{figure}
\centering
\includegraphics[width=\textwidth]{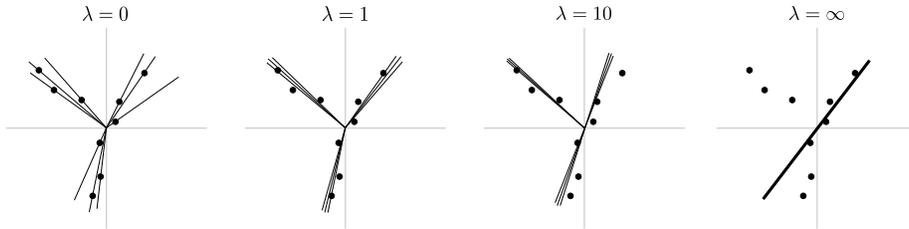}
\caption{
In \eqref{fscEq} and \eqref{ifscEq}, $\lambdaa \geq 0$ regulates how clusters fuse together. If $\lambdaa=0$, each point is assigned to a subspace that exactly contains it (overfitting). The larger $\lambdaa$, the more we penalize subspaces being apart, which results in subspaces getting closer (as with $\lambdaa=1$), up to the point that some subspaces fuse (as with $\lambdaa=10$). In the extreme ($\lambdaa=\infty$), all subspaces fuse together, and we need to explain all data with a single subspace (which may not be enough). Notice that it is not always evident how many subspaces one should use to explain a dataset. In this illustration, should we choose $\lambdaa=1$, which would result in $3$ subspaces, or $\lambdaa=10$, which would result in $2$? Section \ref{modelSelectionSec} discusses how to choose $\lambdaa$, which in turn determines the number of subspaces $\K$ that best explain the data, and their dimensions.}
\label{fscFig}
\end{figure}

In this section we discuss how \FSC\ provides a natural way for model selection, namely determining the number of subspaces $\K$ that best explain the data, and their dimensions.
Intuitively, the first terms in \eqref{fscEq} and \eqref{ifscEq} guarantee that each subspace $\sU_\i$ is close to its assigned column, and the second terms guarantee that subspaces from different columns are close to one another. The tradeoff between these two quantities is determined by $\lambdaa \geq 0$ (see Figure \ref{fscFig}). If $\lambdaa=0$, then the second term is ignored, and there is a trivial solution where each subspace exactly contains its assigned column (thus attaining the minimum, zero, in the first term). If $\lambdaa>0$, the second term forces subspaces from different columns to get closer, even if they no longer contain exactly their assigned columns. As $\lambdaa$ grows, subspaces get closer and closer, up to the point where some subspaces {\em fuse} into one. This is verified in our experiments (see Figure \ref{lambdaKFig}). The extreme case ($\lambdaa=\infty$) forces all subspaces to fuse into one (to attain zero in the second term), meaning we only have one subspace to explain all data, which is precisely \PCA\ (for full data) and \LRMC\ (for incomplete data). In other words, \FSC\ is a generalization of \PCA\ and \LRMC\, which are the particular cases of \eqref{fscEq} and \eqref{ifscEq} with $\lambdaa=\infty$.

This way, $\lambdaa=\infty$ will produce a single subspace cluster. Iteratively decreasing $\lambdaa$ will result in more and more clusters, until $\lambdaa=0$ produces $\n$, analogous to the {\em clusterpath} produced in \cite{clusterpath} for euclidian clustering. The more subspaces, the more accuracy, but the more degrees of freedom (overfitting). For each $\lambdaa$ that provides a different clustering, we can compute a goodness of fit test (like the Akaike information criterion, AIC \cite{akaike}) that quantifies the tradeoff between accuracy and degrees of freedom, to determine the best number of subspaces $\K$. For example, this test can be in the form of $\K$, and the residuals of the projections of each $\x{}_{\i}^\o$ onto its corresponding $\bhat{\U}{}_\k^\o$, as defined in Section \ref{completionSec}. Similarly, we can iteratively increase $\r$ to find all the columns that lie in $1$-dimensional subspaces, then all the columns that lie in $2$-dimensional subspaces, and so on (pruning the data at each iteration).  After this we will have an estimate of the number of subspaces $\K$, and their dimensions.


\section{Experiments}
\label{experimentsSec}
In this section we study the performance of \FSC\ on both, synthetic and real data. For reference, we compare against the following subspace clustering algorithms that allow missing data:
$(a)$ Entry-wise zero-filling followed by \SSC\ (EWZF-SSC) \cite{ewzf}.
$(b)$ Matrix Completion plus \SSC\ (MC+SSC) \cite{ewzf}.
$(c)$ \SSC-Lifting (SSCL) \cite{elhamifar}.
$(c)$ Algebraic variety high-rank matrix completion (AVHRMC) \cite{greg}.
$(e)$ $k$-subspaces with missing data ($k$S) \cite{kGROUSE}.
$(f)$ Expectation-maximization (EM) \cite{ssp14}.
$(g)$ Group-sparse subspace clustering (GSSC) \cite{gssc}.
$(h)$ Mixture subspace clustering (MSC) \cite{gssc}.
We chose these algorithms based on \cite{gssc, elhamifar}, where they show comparable state-of-the-art performance. With full data, $(a)$-$(b)$ simplify to the acclaimed state-of-the-art \SSC. $(c)$-$(d)$ are both lifting schemes. Same as $(a)$-$(b)$, they are two-step procedures that at some point require using \SSC. $(e)$-$(h)$ are alternating algorithms that depend heavily on initialization; according to \cite{gssc}, they produce best results when initialized with the output of 1, and so indirectly they also depend on \SSC. To measure performance we compute clustering error (fraction of misclassified points).

\subsection{Simulations}
Since \FSC\ is an entirely new subspace clustering approach to both, full and incomplete data, we present a thorough series of experiments to study its behavior as a function of the penalty parameter $\lambdaa$, the ambient dimension $\d$, the number of subspaces involved $\K$, their dimensions $\r$, the noise variance \phantomsection\label{sigmaDef}$\sigmaa{}^2$, the number of data points in each cluster $\n_\k$, and of course, the fraction of observed entries \phantomsection\label{pDef}$\p$. Unless otherwise stated, we use the following default settings: $\d=100$, $\K=4$, $\r=5$, $\sigmaa=0$, $\n_\k=20$, and $\p=1$. We run $30$ trials of each experiment, and show the average results of \FSC\, and the best result of each experiment amongst algorithms $(a)$-$(h)$ above.

\sloppypar In all our simulations we first generate $\K$ matrices $\U{}_\k^\star \in \R{}^{\d \times \r}$ with i.i.d.~$\mathscr{N}(0,1)$ entries, to use as bases of the {\em true} subspaces. For each $\k$ we generate a matrix \phantomsection\label{bThetaDef}$\bTheta{}^\star_\k \in \R{}^{\r \times \n_\k}$, also with i.i.d.~$\mathscr{N}(0,1)$ entries, to use as coefficients of the columns in the $\k{}^{\rm th}$ subspace. We then form $\X$ as the concatenation $[\U_1{}^\star\bTheta{}^\star_1 \ \ \ \ \U_2{}^\star\bTheta{}^\star_2 \ \ \cdots \ \ \U{}_\K^\star\bTheta{}^\star_\K]$, plus a $\d\times\n$ noise matrix with i.i.d.~$\mathscr{N}(0,\sigmaa{}^2)$ entries. To create $\O$, we sample each entry independently with probability $\p$. \vspace{.25cm}

\begin{figure}[b]
\centering
\begin{tikzpicture}[x=1pt,y=1pt]
\definecolor{fillColor}{RGB}{255,255,255}
\path[use as bounding box,fill=fillColor,fill opacity=0.00] (0,0) rectangle (144.54,130.09);
\begin{scope}
\path[clip] (  0.00,  0.00) rectangle (144.54,130.09);
\definecolor{drawColor}{RGB}{255,255,255}
\definecolor{fillColor}{RGB}{255,255,255}

\path[draw=drawColor,line width= 0.6pt,line join=round,line cap=round,fill=fillColor] (  0.00,  0.00) rectangle (144.54,130.09);
\end{scope}
\begin{scope}
\path[clip] ( 32.42, 31.48) rectangle (139.04,111.93);
\definecolor{fillColor}{gray}{0.92}

\path[fill=fillColor] ( 32.42, 31.48) rectangle (139.04,111.93);
\definecolor{drawColor}{RGB}{255,255,255}

\path[draw=drawColor,line width= 0.3pt,line join=round] ( 32.42, 40.18) --
	(139.04, 40.18);

\path[draw=drawColor,line width= 0.3pt,line join=round] ( 32.42, 51.53) --
	(139.04, 51.53);

\path[draw=drawColor,line width= 0.3pt,line join=round] ( 32.42, 83.05) --
	(139.04, 83.05);

\path[draw=drawColor,line width= 0.3pt,line join=round] (123.42, 31.48) --
	(123.42,111.93);

\path[draw=drawColor,line width= 0.3pt,line join=round] ( 96.50, 31.48) --
	( 96.50,111.93);

\path[draw=drawColor,line width= 0.3pt,line join=round] ( 69.57, 31.48) --
	( 69.57,111.93);

\path[draw=drawColor,line width= 0.3pt,line join=round] ( 48.03, 31.48) --
	( 48.03,111.93);

\path[draw=drawColor,line width= 0.6pt,line join=round] ( 32.42, 35.14) --
	(139.04, 35.14);

\path[draw=drawColor,line width= 0.6pt,line join=round] ( 32.42, 45.22) --
	(139.04, 45.22);

\path[draw=drawColor,line width= 0.6pt,line join=round] ( 32.42, 57.83) --
	(139.04, 57.83);

\path[draw=drawColor,line width= 0.6pt,line join=round] ( 32.42,108.27) --
	(139.04,108.27);

\path[draw=drawColor,line width= 0.6pt,line join=round] (134.19, 31.48) --
	(134.19,111.93);

\path[draw=drawColor,line width= 0.6pt,line join=round] (112.65, 31.48) --
	(112.65,111.93);

\path[draw=drawColor,line width= 0.6pt,line join=round] ( 80.34, 31.48) --
	( 80.34,111.93);

\path[draw=drawColor,line width= 0.6pt,line join=round] ( 58.80, 31.48) --
	( 58.80,111.93);

\path[draw=drawColor,line width= 0.6pt,line join=round] ( 37.26, 31.48) --
	( 37.26,111.93);
\definecolor{drawColor}{RGB}{0,0,255}

\path[draw=drawColor,line width= 1.7pt,line join=round] ( 37.26,108.27) --
	( 48.03, 57.83) --
	( 58.80, 55.31) --
	( 69.57, 52.79) --
	( 80.34, 47.75) --
	( 91.11, 45.22) --
	(101.88, 42.70) --
	(112.65, 40.18) --
	(123.42, 37.66) --
	(134.19, 35.14);
\end{scope}
\begin{scope}
\path[clip] (  0.00,  0.00) rectangle (144.54,130.09);
\definecolor{drawColor}{RGB}{190,190,190}

\node[text=drawColor,anchor=base,inner sep=0pt, outer sep=0pt, scale=  0.79] at ( 23.51, 32.41) {1};

\node[text=drawColor,anchor=base,inner sep=0pt, outer sep=0pt, scale=  0.79] at ( 23.51, 42.50) {5};

\node[text=drawColor,anchor=base,inner sep=0pt, outer sep=0pt, scale=  0.79] at ( 23.51, 55.11) {10};

\node[text=drawColor,anchor=base,inner sep=0pt, outer sep=0pt, scale=  0.79] at ( 23.51,105.54) {80};
\end{scope}
\begin{scope}
\path[clip] (  0.00,  0.00) rectangle (144.54,130.09);
\definecolor{drawColor}{gray}{0.20}

\path[draw=drawColor,line width= 0.6pt,line join=round] ( 29.67, 35.14) --
	( 32.42, 35.14);

\path[draw=drawColor,line width= 0.6pt,line join=round] ( 29.67, 45.22) --
	( 32.42, 45.22);

\path[draw=drawColor,line width= 0.6pt,line join=round] ( 29.67, 57.83) --
	( 32.42, 57.83);

\path[draw=drawColor,line width= 0.6pt,line join=round] ( 29.67,108.27) --
	( 32.42,108.27);
\end{scope}
\begin{scope}
\path[clip] (  0.00,  0.00) rectangle (144.54,130.09);
\definecolor{drawColor}{gray}{0.20}

\path[draw=drawColor,line width= 0.6pt,line join=round] (134.19, 28.73) --
	(134.19, 31.48);

\path[draw=drawColor,line width= 0.6pt,line join=round] (112.65, 28.73) --
	(112.65, 31.48);

\path[draw=drawColor,line width= 0.6pt,line join=round] ( 80.34, 28.73) --
	( 80.34, 31.48);

\path[draw=drawColor,line width= 0.6pt,line join=round] ( 58.80, 28.73) --
	( 58.80, 31.48);

\path[draw=drawColor,line width= 0.6pt,line join=round] ( 37.26, 28.73) --
	( 37.26, 31.48);
\end{scope}
\begin{scope}
\path[clip] (  0.00,  0.00) rectangle (144.54,130.09);
\definecolor{drawColor}{RGB}{190,190,190}

\node[text=drawColor,anchor=base,inner sep=0pt, outer sep=0pt, scale=  0.79] at (134.19, 21.08) {1};

\node[text=drawColor,anchor=base,inner sep=0pt, outer sep=0pt, scale=  0.79] at (112.65, 21.08) {$10^{-1}$};

\node[text=drawColor,anchor=base,inner sep=0pt, outer sep=0pt, scale=  0.79] at ( 80.34, 21.08) {$10^{-5}$};

\node[text=drawColor,anchor=base,inner sep=0pt, outer sep=0pt, scale=  0.79] at ( 58.80, 21.08) {$10^{-6}$};

\node[text=drawColor,anchor=base,inner sep=0pt, outer sep=0pt, scale=  0.79] at ( 37.26, 21.08) {0};
\end{scope}
\begin{scope}
\path[clip] (  0.00,  0.00) rectangle (144.54,130.09);
\definecolor{drawColor}{RGB}{0,0,0}

\node[text=drawColor,anchor=base,inner sep=0pt, outer sep=0pt, scale=  1.10] at ( 85.73,  8.00) {Parameter $\lambda$};
\end{scope}
\begin{scope}
\path[clip] (  0.00,  0.00) rectangle (144.54,130.09);
\definecolor{drawColor}{RGB}{0,0,0}

\node[text=drawColor,rotate= 90.00,anchor=base,inner sep=0pt, outer sep=0pt, scale=  1.10] at ( 13.08, 71.70) {Produced Clusters};
\end{scope}
\end{tikzpicture}
\caption{Number of clusters obtained by \FSC\ as a function of the parameter $\lambdaa$ in \eqref{fscEq} and \eqref{ifscEq}.}
\label{lambdaKFig}
\end{figure}
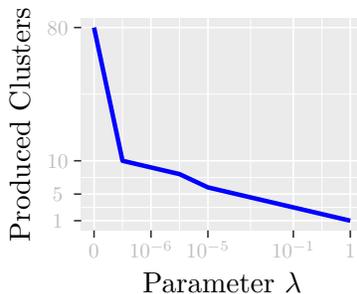

\begin{figure}
\centering
\Scale[.59]{
\begin{tikzpicture}[x=1pt,y=1pt]
\definecolor{fillColor}{RGB}{255,255,255}
\path[use as bounding box,fill=fillColor,fill opacity=0.00] (0,0) rectangle (144.54,130.09);
\begin{scope}
\path[clip] (  0.00,  0.00) rectangle (144.54,130.09);
\definecolor{drawColor}{RGB}{255,255,255}
\definecolor{fillColor}{RGB}{255,255,255}

\path[draw=drawColor,line width= 0.6pt,line join=round,line cap=round,fill=fillColor] ( -0.00,  0.00) rectangle (144.54,130.09);
\end{scope}
\begin{scope}
\path[clip] ( 39.55, 31.48) rectangle (139.04,111.93);
\definecolor{fillColor}{gray}{0.92}

\path[fill=fillColor] ( 39.55, 31.48) rectangle (139.04,111.93);
\definecolor{drawColor}{RGB}{255,255,255}

\path[draw=drawColor,line width= 0.3pt,line join=round] ( 39.55, 41.33) --
	(139.04, 41.33);

\path[draw=drawColor,line width= 0.3pt,line join=round] ( 39.55, 58.06) --
	(139.04, 58.06);

\path[draw=drawColor,line width= 0.3pt,line join=round] ( 39.55, 74.80) --
	(139.04, 74.80);

\path[draw=drawColor,line width= 0.3pt,line join=round] ( 39.55, 91.53) --
	(139.04, 91.53);

\path[draw=drawColor,line width= 0.3pt,line join=round] ( 39.55,108.27) --
	(139.04,108.27);

\path[draw=drawColor,line width= 0.3pt,line join=round] ( 65.97, 31.48) --
	( 65.97,111.93);

\path[draw=drawColor,line width= 0.3pt,line join=round] (111.20, 31.48) --
	(111.20,111.93);

\path[draw=drawColor,line width= 0.6pt,line join=round] ( 39.55, 32.96) --
	(139.04, 32.96);

\path[draw=drawColor,line width= 0.6pt,line join=round] ( 39.55, 49.70) --
	(139.04, 49.70);

\path[draw=drawColor,line width= 0.6pt,line join=round] ( 39.55, 66.43) --
	(139.04, 66.43);

\path[draw=drawColor,line width= 0.6pt,line join=round] ( 39.55, 83.17) --
	(139.04, 83.17);

\path[draw=drawColor,line width= 0.6pt,line join=round] ( 39.55, 99.90) --
	(139.04, 99.90);

\path[draw=drawColor,line width= 0.6pt,line join=round] ( 44.07, 31.48) --
	( 44.07,111.93);

\path[draw=drawColor,line width= 0.6pt,line join=round] ( 87.88, 31.48) --
	( 87.88,111.93);

\path[draw=drawColor,line width= 0.6pt,line join=round] (134.52, 31.48) --
	(134.52,111.93);
\definecolor{drawColor}{RGB}{0,0,255}

\path[draw=drawColor,line width= 1.7pt,line join=round] ( 44.07,108.27) --
	( 46.90, 86.51) --
	( 49.72, 74.80) --
	( 52.55, 56.06) --
	( 55.38, 37.14) --
	( 58.20, 37.14) --
	( 61.03, 37.14) --
	( 63.85, 37.14) --
	( 66.68, 37.14) --
	( 69.51, 35.14) --
	( 72.33, 37.14) --
	( 75.16, 37.14) --
	( 77.99, 37.14) --
	( 80.81, 37.14) --
	( 83.64, 37.14) --
	( 86.47, 37.14) --
	( 89.29, 37.14) --
	( 92.12, 37.14) --
	( 94.95, 37.14) --
	( 97.77, 37.14) --
	(100.60, 37.14) --
	(103.43, 37.14) --
	(106.25, 37.14) --
	(109.08, 37.14) --
	(111.91, 37.14) --
	(114.73, 37.14) --
	(117.56, 56.06) --
	(120.38, 56.06) --
	(123.21, 51.87) --
	(126.04, 51.87) --
	(128.86, 74.80) --
	(131.69, 86.51) --
	(134.52,108.27);
\end{scope}
\begin{scope}
\path[clip] (  0.00,  0.00) rectangle (144.54,130.09);
\definecolor{drawColor}{RGB}{190,190,190}

\node[text=drawColor,anchor=base,inner sep=0pt, outer sep=0pt, scale=  0.79] at ( 27.56, 30.23) {0.00};

\node[text=drawColor,anchor=base,inner sep=0pt, outer sep=0pt, scale=  0.79] at ( 27.56, 46.97) {0.01};

\node[text=drawColor,anchor=base,inner sep=0pt, outer sep=0pt, scale=  0.79] at ( 27.56, 63.70) {0.02};

\node[text=drawColor,anchor=base,inner sep=0pt, outer sep=0pt, scale=  0.79] at ( 27.56, 80.44) {0.03};

\node[text=drawColor,anchor=base,inner sep=0pt, outer sep=0pt, scale=  0.79] at ( 27.56, 97.17) {0.04};
\end{scope}
\begin{scope}
\path[clip] (  0.00,  0.00) rectangle (144.54,130.09);
\definecolor{drawColor}{gray}{0.20}

\path[draw=drawColor,line width= 0.6pt,line join=round] ( 36.80, 32.96) --
	( 39.55, 32.96);

\path[draw=drawColor,line width= 0.6pt,line join=round] ( 36.80, 49.70) --
	( 39.55, 49.70);

\path[draw=drawColor,line width= 0.6pt,line join=round] ( 36.80, 66.43) --
	( 39.55, 66.43);

\path[draw=drawColor,line width= 0.6pt,line join=round] ( 36.80, 83.17) --
	( 39.55, 83.17);

\path[draw=drawColor,line width= 0.6pt,line join=round] ( 36.80, 99.90) --
	( 39.55, 99.90);
\end{scope}
\begin{scope}
\path[clip] (  0.00,  0.00) rectangle (144.54,130.09);
\definecolor{drawColor}{gray}{0.20}

\path[draw=drawColor,line width= 0.6pt,line join=round] ( 44.07, 28.73) --
	( 44.07, 31.48);

\path[draw=drawColor,line width= 0.6pt,line join=round] ( 87.88, 28.73) --
	( 87.88, 31.48);

\path[draw=drawColor,line width= 0.6pt,line join=round] (134.52, 28.73) --
	(134.52, 31.48);
\end{scope}
\begin{scope}
\path[clip] (  0.00,  0.00) rectangle (144.54,130.09);
\definecolor{drawColor}{RGB}{190,190,190}

\node[text=drawColor,anchor=base,inner sep=0pt, outer sep=0pt, scale=  0.79] at ( 44.07, 21.08) {$10^{-5}$};

\node[text=drawColor,anchor=base,inner sep=0pt, outer sep=0pt, scale=  0.79] at ( 87.88, 21.08) {$10^{-3}$};

\node[text=drawColor,anchor=base,inner sep=0pt, outer sep=0pt, scale=  0.79] at (134.52, 21.08) {$10^{-1}$};
\end{scope}
\begin{scope}
\path[clip] (  0.00,  0.00) rectangle (144.54,130.09);
\definecolor{drawColor}{RGB}{0,0,0}

\node[text=drawColor,anchor=base,inner sep=0pt, outer sep=0pt, scale=  1.10] at ( 89.29,  8.00) {Parameter $\lambda$};
\end{scope}
\begin{scope}
\path[clip] (  0.00,  0.00) rectangle (144.54,130.09);
\definecolor{drawColor}{RGB}{0,0,0}

\node[text=drawColor,rotate= 90.00,anchor=base,inner sep=0pt, outer sep=0pt, scale=  1.10] at ( 13.08, 71.70) {Clustering Error};
\end{scope}
\end{tikzpicture}} \hspace{-.2cm}
\Scale[.59]{
\begin{tikzpicture}[x=1pt,y=1pt]
\definecolor{fillColor}{RGB}{255,255,255}
\path[use as bounding box,fill=fillColor,fill opacity=0.00] (0,0) rectangle (144.54,130.09);
\begin{scope}
\path[clip] (  0.00,  0.00) rectangle (144.54,130.09);
\definecolor{drawColor}{RGB}{255,255,255}
\definecolor{fillColor}{RGB}{255,255,255}

\path[draw=drawColor,line width= 0.6pt,line join=round,line cap=round,fill=fillColor] (  0.00,  0.00) rectangle (144.54,130.09);
\end{scope}
\begin{scope}
\path[clip] ( 33.64, 29.95) rectangle (139.04,111.93);
\definecolor{fillColor}{gray}{0.92}

\path[fill=fillColor] ( 33.64, 29.95) rectangle (139.04,111.93);
\definecolor{drawColor}{RGB}{255,255,255}

\path[draw=drawColor,line width= 0.3pt,line join=round] ( 33.64, 48.58) --
	(139.04, 48.58);

\path[draw=drawColor,line width= 0.3pt,line join=round] ( 33.64, 72.43) --
	(139.04, 72.43);

\path[draw=drawColor,line width= 0.3pt,line join=round] ( 33.64, 96.28) --
	(139.04, 96.28);

\path[draw=drawColor,line width= 0.3pt,line join=round] ( 54.40, 29.95) --
	( 54.40,111.93);

\path[draw=drawColor,line width= 0.3pt,line join=round] ( 86.34, 29.95) --
	( 86.34,111.93);

\path[draw=drawColor,line width= 0.3pt,line join=round] (118.28, 29.95) --
	(118.28,111.93);

\path[draw=drawColor,line width= 0.6pt,line join=round] ( 33.64, 36.66) --
	(139.04, 36.66);

\path[draw=drawColor,line width= 0.6pt,line join=round] ( 33.64, 60.51) --
	(139.04, 60.51);

\path[draw=drawColor,line width= 0.6pt,line join=round] ( 33.64, 84.35) --
	(139.04, 84.35);

\path[draw=drawColor,line width= 0.6pt,line join=round] ( 33.64,108.20) --
	(139.04,108.20);

\path[draw=drawColor,line width= 0.6pt,line join=round] ( 38.43, 29.95) --
	( 38.43,111.93);

\path[draw=drawColor,line width= 0.6pt,line join=round] ( 70.37, 29.95) --
	( 70.37,111.93);

\path[draw=drawColor,line width= 0.6pt,line join=round] (102.31, 29.95) --
	(102.31,111.93);

\path[draw=drawColor,line width= 0.6pt,line join=round] (134.25, 29.95) --
	(134.25,111.93);
\definecolor{drawColor}{RGB}{0,0,0}

\path[draw=drawColor,line width= 1.7pt,line join=round] ( 38.43, 36.66) --
	( 54.40, 36.66) --
	( 70.37, 36.66) --
	( 86.34, 36.66) --
	(102.31, 36.66) --
	(118.28, 97.33) --
	(134.25,106.86);
\definecolor{drawColor}{RGB}{0,0,255}

\path[draw=drawColor,line width= 1.7pt,line join=round] ( 38.43, 36.81) --
	( 54.40, 36.81) --
	( 70.37, 36.96) --
	( 86.34, 36.66) --
	(102.31, 36.66) --
	(118.28, 37.41) --
	(134.25, 43.81);
\end{scope}
\begin{scope}
\path[clip] (  0.00,  0.00) rectangle (144.54,130.09);
\definecolor{drawColor}{RGB}{190,190,190}

\node[text=drawColor,anchor=base,inner sep=0pt, outer sep=0pt, scale=  0.79] at ( 23.63, 33.93) {0.0};

\node[text=drawColor,anchor=base,inner sep=0pt, outer sep=0pt, scale=  0.79] at ( 23.63, 57.78) {0.2};

\node[text=drawColor,anchor=base,inner sep=0pt, outer sep=0pt, scale=  0.79] at ( 23.63, 81.63) {0.4};

\node[text=drawColor,anchor=base,inner sep=0pt, outer sep=0pt, scale=  0.79] at ( 23.63,105.47) {0.6};
\end{scope}
\begin{scope}
\path[clip] (  0.00,  0.00) rectangle (144.54,130.09);
\definecolor{drawColor}{gray}{0.20}

\path[draw=drawColor,line width= 0.6pt,line join=round] ( 30.89, 36.66) --
	( 33.64, 36.66);

\path[draw=drawColor,line width= 0.6pt,line join=round] ( 30.89, 60.51) --
	( 33.64, 60.51);

\path[draw=drawColor,line width= 0.6pt,line join=round] ( 30.89, 84.35) --
	( 33.64, 84.35);

\path[draw=drawColor,line width= 0.6pt,line join=round] ( 30.89,108.20) --
	( 33.64,108.20);
\end{scope}
\begin{scope}
\path[clip] (  0.00,  0.00) rectangle (144.54,130.09);
\definecolor{drawColor}{gray}{0.20}

\path[draw=drawColor,line width= 0.6pt,line join=round] ( 38.43, 27.20) --
	( 38.43, 29.95);

\path[draw=drawColor,line width= 0.6pt,line join=round] ( 70.37, 27.20) --
	( 70.37, 29.95);

\path[draw=drawColor,line width= 0.6pt,line join=round] (102.31, 27.20) --
	(102.31, 29.95);

\path[draw=drawColor,line width= 0.6pt,line join=round] (134.25, 27.20) --
	(134.25, 29.95);
\end{scope}
\begin{scope}
\path[clip] (  0.00,  0.00) rectangle (144.54,130.09);
\definecolor{drawColor}{RGB}{190,190,190}

\node[text=drawColor,anchor=base,inner sep=0pt, outer sep=0pt, scale=  0.79] at ( 38.43, 19.55) {$10^{-4}$};

\node[text=drawColor,anchor=base,inner sep=0pt, outer sep=0pt, scale=  0.79] at ( 70.37, 19.55) {$10^{-3}$};

\node[text=drawColor,anchor=base,inner sep=0pt, outer sep=0pt, scale=  0.79] at (102.31, 19.55) {$10^{-2}$};

\node[text=drawColor,anchor=base,inner sep=0pt, outer sep=0pt, scale=  0.79] at (134.25, 19.55) {$10^{-1}$};
\end{scope}
\begin{scope}
\path[clip] (  0.00,  0.00) rectangle (144.54,130.09);
\definecolor{drawColor}{RGB}{0,0,0}

\node[text=drawColor,anchor=base,inner sep=0pt, outer sep=0pt, scale=  1.10] at ( 86.34,  6.47) {Noise Level};
\end{scope}
\begin{scope}
\path[clip] (  0.00,  0.00) rectangle (144.54,130.09);

\path[] ( 34.52, 73.89) rectangle (126.36,125.37);
\end{scope}
\begin{scope}
\path[clip] (  0.00,  0.00) rectangle (144.54,130.09);

\path[] ( 40.21, 94.03) rectangle ( 54.66,108.49);
\end{scope}
\begin{scope}
\path[clip] (  0.00,  0.00) rectangle (144.54,130.09);
\definecolor{drawColor}{RGB}{0,0,255}

\path[draw=drawColor,line width= 1.7pt,line join=round] ( 41.65,101.26) -- ( 53.22,101.26);
\end{scope}
\begin{scope}
\path[clip] (  0.00,  0.00) rectangle (144.54,130.09);
\definecolor{drawColor}{RGB}{0,0,255}

\path[draw=drawColor,line width= 1.7pt,line join=round] ( 41.65,101.26) -- ( 53.22,101.26);
\end{scope}
\begin{scope}
\path[clip] (  0.00,  0.00) rectangle (144.54,130.09);

\path[] ( 40.21, 79.58) rectangle ( 54.66, 94.03);
\end{scope}
\begin{scope}
\path[clip] (  0.00,  0.00) rectangle (144.54,130.09);
\definecolor{drawColor}{RGB}{0,0,0}

\path[draw=drawColor,line width= 1.7pt,line join=round] ( 41.65, 86.81) -- ( 53.22, 86.81);
\end{scope}
\begin{scope}
\path[clip] (  0.00,  0.00) rectangle (144.54,130.09);
\definecolor{drawColor}{RGB}{0,0,0}

\path[draw=drawColor,line width= 1.7pt,line join=round] ( 41.65, 86.81) -- ( 53.22, 86.81);
\end{scope}
\begin{scope}
\path[clip] (  0.00,  0.00) rectangle (144.54,130.09);
\definecolor{drawColor}{RGB}{0,0,0}

\node[text=drawColor,anchor=base west,inner sep=0pt, outer sep=0pt, scale=  0.88] at ( 56.47, 98.23) {{\sc Fsc} (this paper)};
\end{scope}
\begin{scope}
\path[clip] (  0.00,  0.00) rectangle (144.54,130.09);
\definecolor{drawColor}{RGB}{0,0,0}

\node[text=drawColor,anchor=base west,inner sep=0pt, outer sep=0pt, scale=  0.88] at ( 56.47, 83.78) {Best$(a$-$h)$};
\end{scope}
\end{tikzpicture}} \hspace{-.2cm}
\Scale[.59]{
\begin{tikzpicture}[x=1pt,y=1pt]
\definecolor{fillColor}{RGB}{255,255,255}
\path[use as bounding box,fill=fillColor,fill opacity=0.00] (0,0) rectangle (144.54,130.09);
\begin{scope}
\path[clip] (  0.00,  0.00) rectangle (144.54,130.09);
\definecolor{drawColor}{RGB}{255,255,255}
\definecolor{fillColor}{RGB}{255,255,255}

\path[draw=drawColor,line width= 0.6pt,line join=round,line cap=round,fill=fillColor] ( -0.00,  0.00) rectangle (144.54,130.09);
\end{scope}
\begin{scope}
\path[clip] ( 37.60, 29.95) rectangle (139.04,111.93);
\definecolor{fillColor}{gray}{0.92}

\path[fill=fillColor] ( 37.60, 29.95) rectangle (139.04,111.93);
\definecolor{drawColor}{RGB}{255,255,255}

\path[draw=drawColor,line width= 0.3pt,line join=round] ( 37.60, 44.66) --
	(139.04, 44.66);

\path[draw=drawColor,line width= 0.3pt,line join=round] ( 37.60, 66.62) --
	(139.04, 66.62);

\path[draw=drawColor,line width= 0.3pt,line join=round] ( 37.60, 88.58) --
	(139.04, 88.58);

\path[draw=drawColor,line width= 0.3pt,line join=round] ( 37.60,110.54) --
	(139.04,110.54);

\path[draw=drawColor,line width= 0.3pt,line join=round] ( 37.60, 29.95) --
	( 37.60,111.93);

\path[draw=drawColor,line width= 0.3pt,line join=round] ( 46.82, 29.95) --
	( 46.82,111.93);

\path[draw=drawColor,line width= 0.3pt,line join=round] ( 60.66, 29.95) --
	( 60.66,111.93);

\path[draw=drawColor,line width= 0.3pt,line join=round] ( 88.32, 29.95) --
	( 88.32,111.93);

\path[draw=drawColor,line width= 0.3pt,line join=round] (120.60, 29.95) --
	(120.60,111.93);

\path[draw=drawColor,line width= 0.3pt,line join=round] (139.04, 29.95) --
	(139.04,111.93);

\path[draw=drawColor,line width= 0.6pt,line join=round] ( 37.60, 33.68) --
	(139.04, 33.68);

\path[draw=drawColor,line width= 0.6pt,line join=round] ( 37.60, 55.64) --
	(139.04, 55.64);

\path[draw=drawColor,line width= 0.6pt,line join=round] ( 37.60, 77.60) --
	(139.04, 77.60);

\path[draw=drawColor,line width= 0.6pt,line join=round] ( 37.60, 99.56) --
	(139.04, 99.56);

\path[draw=drawColor,line width= 0.6pt,line join=round] ( 42.21, 29.95) --
	( 42.21,111.93);

\path[draw=drawColor,line width= 0.6pt,line join=round] ( 51.44, 29.95) --
	( 51.44,111.93);

\path[draw=drawColor,line width= 0.6pt,line join=round] ( 69.88, 29.95) --
	( 69.88,111.93);

\path[draw=drawColor,line width= 0.6pt,line join=round] (106.76, 29.95) --
	(106.76,111.93);

\path[draw=drawColor,line width= 0.6pt,line join=round] (134.43, 29.95) --
	(134.43,111.93);
\definecolor{drawColor}{RGB}{0,0,0}

\path[draw=drawColor,line width= 1.7pt,line join=round] ( 42.21, 99.98) --
	( 51.44, 42.46) --
	( 60.66, 33.68) --
	( 69.88, 33.68) --
	( 79.10, 33.68) --
	( 88.32, 33.68) --
	( 97.54, 33.68) --
	(106.76, 33.68) --
	(115.99, 33.68) --
	(125.21, 33.68) --
	(134.43, 33.68);
\definecolor{drawColor}{RGB}{0,0,255}

\path[draw=drawColor,line width= 1.7pt,line join=round] ( 42.21,108.20) --
	( 51.44, 78.70) --
	( 60.66, 79.79) --
	( 69.88, 78.70) --
	( 79.10, 52.34) --
	( 88.32, 41.69) --
	( 97.54, 39.17) --
	(106.76, 36.42) --
	(115.99, 34.78) --
	(125.21, 34.23) --
	(134.43, 33.95);
\end{scope}
\begin{scope}
\path[clip] (  0.00,  0.00) rectangle (144.54,130.09);
\definecolor{drawColor}{RGB}{190,190,190}

\node[text=drawColor,anchor=base,inner sep=0pt, outer sep=0pt, scale=  0.79] at ( 25.61, 30.95) {0.00};

\node[text=drawColor,anchor=base,inner sep=0pt, outer sep=0pt, scale=  0.79] at ( 25.61, 52.91) {0.02};

\node[text=drawColor,anchor=base,inner sep=0pt, outer sep=0pt, scale=  0.79] at ( 25.61, 74.87) {0.04};

\node[text=drawColor,anchor=base,inner sep=0pt, outer sep=0pt, scale=  0.79] at ( 25.61, 96.83) {0.06};
\end{scope}
\begin{scope}
\path[clip] (  0.00,  0.00) rectangle (144.54,130.09);
\definecolor{drawColor}{gray}{0.20}

\path[draw=drawColor,line width= 0.6pt,line join=round] ( 34.85, 33.68) --
	( 37.60, 33.68);

\path[draw=drawColor,line width= 0.6pt,line join=round] ( 34.85, 55.64) --
	( 37.60, 55.64);

\path[draw=drawColor,line width= 0.6pt,line join=round] ( 34.85, 77.60) --
	( 37.60, 77.60);

\path[draw=drawColor,line width= 0.6pt,line join=round] ( 34.85, 99.56) --
	( 37.60, 99.56);
\end{scope}
\begin{scope}
\path[clip] (  0.00,  0.00) rectangle (144.54,130.09);
\definecolor{drawColor}{gray}{0.20}

\path[draw=drawColor,line width= 0.6pt,line join=round] ( 42.21, 27.20) --
	( 42.21, 29.95);

\path[draw=drawColor,line width= 0.6pt,line join=round] ( 51.44, 27.20) --
	( 51.44, 29.95);

\path[draw=drawColor,line width= 0.6pt,line join=round] ( 69.88, 27.20) --
	( 69.88, 29.95);

\path[draw=drawColor,line width= 0.6pt,line join=round] (106.76, 27.20) --
	(106.76, 29.95);

\path[draw=drawColor,line width= 0.6pt,line join=round] (134.43, 27.20) --
	(134.43, 29.95);
\end{scope}
\begin{scope}
\path[clip] (  0.00,  0.00) rectangle (144.54,130.09);
\definecolor{drawColor}{RGB}{190,190,190}

\node[text=drawColor,anchor=base,inner sep=0pt, outer sep=0pt, scale=  0.79] at ( 42.21, 19.55) {6};

\node[text=drawColor,anchor=base,inner sep=0pt, outer sep=0pt, scale=  0.79] at ( 51.44, 19.55) {10};

\node[text=drawColor,anchor=base,inner sep=0pt, outer sep=0pt, scale=  0.79] at ( 69.88, 19.55) {30};

\node[text=drawColor,anchor=base,inner sep=0pt, outer sep=0pt, scale=  0.79] at (106.76, 19.55) {70};

\node[text=drawColor,anchor=base,inner sep=0pt, outer sep=0pt, scale=  0.79] at (134.43, 19.55) {100};
\end{scope}
\begin{scope}
\path[clip] (  0.00,  0.00) rectangle (144.54,130.09);
\definecolor{drawColor}{RGB}{0,0,0}

\node[text=drawColor,anchor=base,inner sep=0pt, outer sep=0pt, scale=  1.10] at ( 88.32,  6.47) {Ambient Dimension};
\end{scope}
\end{tikzpicture}} \hspace{-.2cm}
\Scale[.59]{
\begin{tikzpicture}[x=1pt,y=1pt]
\definecolor{fillColor}{RGB}{255,255,255}
\path[use as bounding box,fill=fillColor,fill opacity=0.00] (0,0) rectangle (144.54,130.09);
\begin{scope}
\path[clip] (  0.00,  0.00) rectangle (144.54,130.09);
\definecolor{drawColor}{RGB}{255,255,255}
\definecolor{fillColor}{RGB}{255,255,255}

\path[draw=drawColor,line width= 0.6pt,line join=round,line cap=round,fill=fillColor] ( -0.00, -0.00) rectangle (144.54,130.09);
\end{scope}
\begin{scope}
\path[clip] ( 37.60, 30.92) rectangle (139.04,111.93);
\definecolor{fillColor}{gray}{0.92}

\path[fill=fillColor] ( 37.60, 30.92) rectangle (139.04,111.93);
\definecolor{drawColor}{RGB}{255,255,255}

\path[draw=drawColor,line width= 0.3pt,line join=round] ( 37.60, 45.52) --
	(139.04, 45.52);

\path[draw=drawColor,line width= 0.3pt,line join=round] ( 37.60, 67.33) --
	(139.04, 67.33);

\path[draw=drawColor,line width= 0.3pt,line join=round] ( 37.60, 89.15) --
	(139.04, 89.15);

\path[draw=drawColor,line width= 0.3pt,line join=round] ( 37.60,110.97) --
	(139.04,110.97);

\path[draw=drawColor,line width= 0.3pt,line join=round] ( 51.44, 30.92) --
	( 51.44,111.93);

\path[draw=drawColor,line width= 0.3pt,line join=round] ( 69.88, 30.92) --
	( 69.88,111.93);

\path[draw=drawColor,line width= 0.3pt,line join=round] ( 88.32, 30.92) --
	( 88.32,111.93);

\path[draw=drawColor,line width= 0.3pt,line join=round] (106.76, 30.92) --
	(106.76,111.93);

\path[draw=drawColor,line width= 0.3pt,line join=round] (125.21, 30.92) --
	(125.21,111.93);

\path[draw=drawColor,line width= 0.6pt,line join=round] ( 37.60, 34.61) --
	(139.04, 34.61);

\path[draw=drawColor,line width= 0.6pt,line join=round] ( 37.60, 56.42) --
	(139.04, 56.42);

\path[draw=drawColor,line width= 0.6pt,line join=round] ( 37.60, 78.24) --
	(139.04, 78.24);

\path[draw=drawColor,line width= 0.6pt,line join=round] ( 37.60,100.06) --
	(139.04,100.06);

\path[draw=drawColor,line width= 0.6pt,line join=round] ( 42.21, 30.92) --
	( 42.21,111.93);

\path[draw=drawColor,line width= 0.6pt,line join=round] ( 60.66, 30.92) --
	( 60.66,111.93);

\path[draw=drawColor,line width= 0.6pt,line join=round] ( 79.10, 30.92) --
	( 79.10,111.93);

\path[draw=drawColor,line width= 0.6pt,line join=round] ( 97.54, 30.92) --
	( 97.54,111.93);

\path[draw=drawColor,line width= 0.6pt,line join=round] (115.99, 30.92) --
	(115.99,111.93);

\path[draw=drawColor,line width= 0.6pt,line join=round] (134.43, 30.92) --
	(134.43,111.93);
\definecolor{drawColor}{RGB}{0,0,0}

\path[draw=drawColor,line width= 1.7pt,line join=round] ( 42.21, 34.61) --
	( 60.66, 34.61) --
	( 79.10, 34.61) --
	( 97.54, 35.35) --
	(115.99, 86.43) --
	(134.43,106.88);
\definecolor{drawColor}{RGB}{0,0,255}

\path[draw=drawColor,line width= 1.7pt,line join=round] ( 42.21, 35.38) --
	( 60.66, 34.88) --
	( 79.10, 59.70) --
	( 97.54, 95.70) --
	(115.99,105.52) --
	(134.43,108.24);
\end{scope}
\begin{scope}
\path[clip] (  0.00,  0.00) rectangle (144.54,130.09);
\definecolor{drawColor}{RGB}{190,190,190}

\node[text=drawColor,anchor=base,inner sep=0pt, outer sep=0pt, scale=  0.79] at ( 25.61, 31.88) {0.00};

\node[text=drawColor,anchor=base,inner sep=0pt, outer sep=0pt, scale=  0.79] at ( 25.61, 53.70) {0.02};

\node[text=drawColor,anchor=base,inner sep=0pt, outer sep=0pt, scale=  0.79] at ( 25.61, 75.52) {0.04};

\node[text=drawColor,anchor=base,inner sep=0pt, outer sep=0pt, scale=  0.79] at ( 25.61, 97.33) {0.06};
\end{scope}
\begin{scope}
\path[clip] (  0.00,  0.00) rectangle (144.54,130.09);
\definecolor{drawColor}{gray}{0.20}

\path[draw=drawColor,line width= 0.6pt,line join=round] ( 34.85, 34.61) --
	( 37.60, 34.61);

\path[draw=drawColor,line width= 0.6pt,line join=round] ( 34.85, 56.42) --
	( 37.60, 56.42);

\path[draw=drawColor,line width= 0.6pt,line join=round] ( 34.85, 78.24) --
	( 37.60, 78.24);

\path[draw=drawColor,line width= 0.6pt,line join=round] ( 34.85,100.06) --
	( 37.60,100.06);
\end{scope}
\begin{scope}
\path[clip] (  0.00,  0.00) rectangle (144.54,130.09);
\definecolor{drawColor}{gray}{0.20}

\path[draw=drawColor,line width= 0.6pt,line join=round] ( 42.21, 28.17) --
	( 42.21, 30.92);

\path[draw=drawColor,line width= 0.6pt,line join=round] ( 60.66, 28.17) --
	( 60.66, 30.92);

\path[draw=drawColor,line width= 0.6pt,line join=round] ( 79.10, 28.17) --
	( 79.10, 30.92);

\path[draw=drawColor,line width= 0.6pt,line join=round] ( 97.54, 28.17) --
	( 97.54, 30.92);

\path[draw=drawColor,line width= 0.6pt,line join=round] (115.99, 28.17) --
	(115.99, 30.92);

\path[draw=drawColor,line width= 0.6pt,line join=round] (134.43, 28.17) --
	(134.43, 30.92);
\end{scope}
\begin{scope}
\path[clip] (  0.00,  0.00) rectangle (144.54,130.09);
\definecolor{drawColor}{RGB}{190,190,190}

\node[text=drawColor,anchor=base,inner sep=0pt, outer sep=0pt, scale=  0.79] at ( 42.21, 20.52) {1};

\node[text=drawColor,anchor=base,inner sep=0pt, outer sep=0pt, scale=  0.79] at ( 60.66, 20.52) {5};

\node[text=drawColor,anchor=base,inner sep=0pt, outer sep=0pt, scale=  0.79] at ( 79.10, 20.52) {10};

\node[text=drawColor,anchor=base,inner sep=0pt, outer sep=0pt, scale=  0.79] at ( 97.54, 20.52) {20};

\node[text=drawColor,anchor=base,inner sep=0pt, outer sep=0pt, scale=  0.79] at (115.99, 20.52) {30};

\node[text=drawColor,anchor=base,inner sep=0pt, outer sep=0pt, scale=  0.79] at (134.43, 20.52) {50};
\end{scope}
\begin{scope}
\path[clip] (  0.00,  0.00) rectangle (144.54,130.09);
\definecolor{drawColor}{RGB}{0,0,0}

\node[text=drawColor,anchor=base,inner sep=0pt, outer sep=0pt, scale=  1.10] at ( 88.32,  7.44) {Subspaces' Dimension};
\end{scope}
\end{tikzpicture}}
\caption{Clustering error of \FSC\ and the best algorithm among $(a)$-$(h)$ in each trial. Notice the different scales: even in the worst-case (subspace's dimension $\r=20$), the gap between \FSC\ and the best algorithm (on each trial) among $(a)$-$(h)$ is less than $6\%$.}
\label{results1Fig}
\end{figure}
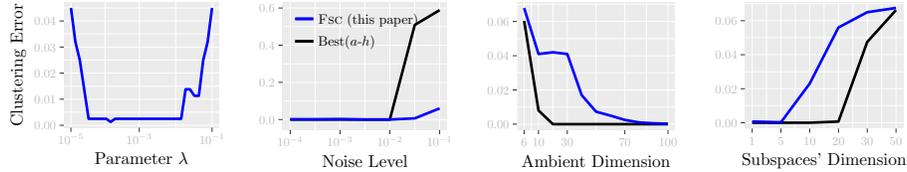

{\bf Effect of the penalty parameter.}
In our first experiment we study the number of clusters obtained by \FSC\ as a function of $\lambdaa$, with the default settings above. Figure \ref{lambdaKFig} shows, consistent with our discussion in Section \ref{modelSelectionSec}, that if $\lambdaa=0$, \FSC\ assigns each point to its own cluster. As $\lambdaa$ increases, subspaces start fusing together up to the point where if $\lambdaa$ is too large, \FSC\ fuses all subspaces into one, and all data gets clustered together. Next we study performance. Figure \ref{results1Fig} shows that there is a wide range of values of $\lambdaa$ that produce low error, which shows that \FSC\ is quite stable. Notice that the error increases if $\lambdaa$ is too small or too large. This is consistent with our previous experiment, showing that this is because such extreme values of $\lambdaa$ produce too few or too many clusters.

{\bf Effect of noise.} Figure \ref{results1Fig} shows that \FSC\ performs as well as the state-of-the-art with low-noise, but considerably better in the high-noise regime. Recall that $\lambdaa$ quantifies the tradeoff between how accurately we want to represent each point $\x_\i$ (the first term in \eqref{fscEq} and \eqref{ifscEq}), and how close subspaces from different points will be (second term), which in turn determines how subspaces fuse together, or equivalently, how many subspaces we will obtain. If data is completely noiseless, we expect to represent each point very accurately, and so we can use a smaller $\lambdaa$ (giving more weight to the first term). On the other hand, if data is noisy, we expect to represent each point within the noise level, and so we can use a larger $\lambdaa$. As a rule of thumb, we can use $\lambdaa$ inversely proportional to the noise level $\sigmaa$.

{\bf Effect of dimensionality.} It is well-documented that data in lower-dimensional subspaces are easier to cluster \cite{infoTheoretic, hrmc, gssc, elhamifar, greg, ladmc}. In the extreme case, clustering $1$-dimensional subspaces requires a simple co-linearity test, and is theoretically possible with as little as $2$ samples per column \cite{infoTheoretic}. In contrast, no existing algorithm can successfully cluster $(\d-1)$-dimensional subspaces (hyperplanes), which is actually impossible even if one entry per column is missing \cite{infoTheoretic}. Of course, being low-dimensional is relative to the ambient dimension: a $10$-dimensional subspace is a hyperplane in $\R{}^{11}$, but low-dimensional in $\R{}^{1000}$. In this experiment we test \FSC\ as a function of the {\em low-dimensionality} of the subspaces, i.e., the gap between the ambient dimension $\d$ and the subspaces' dimension $\r$. First we fix $\r=5$, and compute error as a function of $\d$. As $\d$ grows, this subspace becomes lower and lower-dimensional. Then we turn things around, fixing $\d=100$ and varying $\r$. As $\r$ grows, the subspace becomes higher and higher-dimensional. The results are in Figure \ref{results1Fig}. Unfortunately, \FSC\ seems more sensitive to high-dimensionality than the state-of-the-art. However, pay attention to the scale: even in the worst-case ($\r=20$), the gap between \FSC\ and the best algorithm (on each trial) among $(a)$-$(h)$ is less than $6\%$.

\begin{figure}
\centering
\Scale[.6]{
\begin{tikzpicture}[x=1pt,y=1pt]
\definecolor{fillColor}{RGB}{255,255,255}
\path[use as bounding box,fill=fillColor,fill opacity=0.00] (0,0) rectangle (144.54,130.09);
\begin{scope}
\path[clip] (  0.00,  0.00) rectangle (144.54,130.09);
\definecolor{drawColor}{RGB}{255,255,255}
\definecolor{fillColor}{RGB}{255,255,255}

\path[draw=drawColor,line width= 0.6pt,line join=round,line cap=round,fill=fillColor] ( -0.00, -0.00) rectangle (144.54,130.09);
\end{scope}
\begin{scope}
\path[clip] ( 39.55, 30.92) rectangle (139.04,111.93);
\definecolor{fillColor}{gray}{0.92}

\path[fill=fillColor] ( 39.55, 30.92) rectangle (139.04,111.93);
\definecolor{drawColor}{RGB}{255,255,255}

\path[draw=drawColor,line width= 0.3pt,line join=round] ( 39.55, 37.55) --
	(139.04, 37.55);

\path[draw=drawColor,line width= 0.3pt,line join=round] ( 39.55, 61.12) --
	(139.04, 61.12);

\path[draw=drawColor,line width= 0.3pt,line join=round] ( 39.55, 84.68) --
	(139.04, 84.68);

\path[draw=drawColor,line width= 0.3pt,line join=round] ( 39.55,108.24) --
	(139.04,108.24);

\path[draw=drawColor,line width= 0.3pt,line join=round] ( 51.61, 30.92) --
	( 51.61,111.93);

\path[draw=drawColor,line width= 0.3pt,line join=round] ( 66.68, 30.92) --
	( 66.68,111.93);

\path[draw=drawColor,line width= 0.3pt,line join=round] ( 81.76, 30.92) --
	( 81.76,111.93);

\path[draw=drawColor,line width= 0.3pt,line join=round] ( 96.83, 30.92) --
	( 96.83,111.93);

\path[draw=drawColor,line width= 0.3pt,line join=round] (111.91, 30.92) --
	(111.91,111.93);

\path[draw=drawColor,line width= 0.3pt,line join=round] (126.98, 30.92) --
	(126.98,111.93);

\path[draw=drawColor,line width= 0.6pt,line join=round] ( 39.55, 49.33) --
	(139.04, 49.33);

\path[draw=drawColor,line width= 0.6pt,line join=round] ( 39.55, 72.90) --
	(139.04, 72.90);

\path[draw=drawColor,line width= 0.6pt,line join=round] ( 39.55, 96.46) --
	(139.04, 96.46);

\path[draw=drawColor,line width= 0.6pt,line join=round] ( 44.07, 30.92) --
	( 44.07,111.93);

\path[draw=drawColor,line width= 0.6pt,line join=round] ( 59.14, 30.92) --
	( 59.14,111.93);

\path[draw=drawColor,line width= 0.6pt,line join=round] ( 74.22, 30.92) --
	( 74.22,111.93);

\path[draw=drawColor,line width= 0.6pt,line join=round] ( 89.29, 30.92) --
	( 89.29,111.93);

\path[draw=drawColor,line width= 0.6pt,line join=round] (104.37, 30.92) --
	(104.37,111.93);

\path[draw=drawColor,line width= 0.6pt,line join=round] (119.44, 30.92) --
	(119.44,111.93);

\path[draw=drawColor,line width= 0.6pt,line join=round] (134.52, 30.92) --
	(134.52,111.93);
\definecolor{drawColor}{RGB}{0,0,0}

\path[draw=drawColor,line width= 1.7pt,line join=round] ( 44.07, 49.33) --
	( 59.14, 49.33) --
	( 74.22, 49.33) --
	( 89.29, 49.33) --
	(104.37, 49.33) --
	(119.44, 49.33) --
	(134.52, 49.33);
\definecolor{drawColor}{RGB}{0,0,255}

\path[draw=drawColor,line width= 1.7pt,line join=round] ( 44.07, 49.33) --
	( 59.14, 50.22) --
	( 74.22, 50.34) --
	( 89.29, 50.81) --
	(104.37, 51.22) --
	(119.44, 51.87) --
	(134.52, 51.69);
\end{scope}
\begin{scope}
\path[clip] (  0.00,  0.00) rectangle (144.54,130.09);
\definecolor{drawColor}{RGB}{190,190,190}

\node[text=drawColor,anchor=base,inner sep=0pt, outer sep=0pt, scale=  0.79] at ( 27.56, 46.61) {0.00};

\node[text=drawColor,anchor=base,inner sep=0pt, outer sep=0pt, scale=  0.79] at ( 27.56, 70.17) {0.04};

\node[text=drawColor,anchor=base,inner sep=0pt, outer sep=0pt, scale=  0.79] at ( 27.56, 93.73) {0.08};
\end{scope}
\begin{scope}
\path[clip] (  0.00,  0.00) rectangle (144.54,130.09);
\definecolor{drawColor}{gray}{0.20}

\path[draw=drawColor,line width= 0.6pt,line join=round] ( 36.80, 49.33) --
	( 39.55, 49.33);

\path[draw=drawColor,line width= 0.6pt,line join=round] ( 36.80, 72.90) --
	( 39.55, 72.90);

\path[draw=drawColor,line width= 0.6pt,line join=round] ( 36.80, 96.46) --
	( 39.55, 96.46);
\end{scope}
\begin{scope}
\path[clip] (  0.00,  0.00) rectangle (144.54,130.09);
\definecolor{drawColor}{gray}{0.20}

\path[draw=drawColor,line width= 0.6pt,line join=round] ( 44.07, 28.17) --
	( 44.07, 30.92);

\path[draw=drawColor,line width= 0.6pt,line join=round] ( 59.14, 28.17) --
	( 59.14, 30.92);

\path[draw=drawColor,line width= 0.6pt,line join=round] ( 74.22, 28.17) --
	( 74.22, 30.92);

\path[draw=drawColor,line width= 0.6pt,line join=round] ( 89.29, 28.17) --
	( 89.29, 30.92);

\path[draw=drawColor,line width= 0.6pt,line join=round] (104.37, 28.17) --
	(104.37, 30.92);

\path[draw=drawColor,line width= 0.6pt,line join=round] (119.44, 28.17) --
	(119.44, 30.92);

\path[draw=drawColor,line width= 0.6pt,line join=round] (134.52, 28.17) --
	(134.52, 30.92);
\end{scope}
\begin{scope}
\path[clip] (  0.00,  0.00) rectangle (144.54,130.09);
\definecolor{drawColor}{RGB}{190,190,190}

\node[text=drawColor,anchor=base,inner sep=0pt, outer sep=0pt, scale=  0.79] at ( 44.07, 20.52) {2};

\node[text=drawColor,anchor=base,inner sep=0pt, outer sep=0pt, scale=  0.79] at ( 59.14, 20.52) {5};

\node[text=drawColor,anchor=base,inner sep=0pt, outer sep=0pt, scale=  0.79] at ( 74.22, 20.52) {10};

\node[text=drawColor,anchor=base,inner sep=0pt, outer sep=0pt, scale=  0.79] at ( 89.29, 20.52) {15};

\node[text=drawColor,anchor=base,inner sep=0pt, outer sep=0pt, scale=  0.79] at (104.37, 20.52) {20};

\node[text=drawColor,anchor=base,inner sep=0pt, outer sep=0pt, scale=  0.79] at (119.44, 20.52) {25};

\node[text=drawColor,anchor=base,inner sep=0pt, outer sep=0pt, scale=  0.79] at (134.52, 20.52) {30};
\end{scope}
\begin{scope}
\path[clip] (  0.00,  0.00) rectangle (144.54,130.09);
\definecolor{drawColor}{RGB}{0,0,0}

\node[text=drawColor,anchor=base,inner sep=0pt, outer sep=0pt, scale=  1.10] at ( 89.29,  7.44) {Number of Subspaces};
\end{scope}
\begin{scope}
\path[clip] (  0.00,  0.00) rectangle (144.54,130.09);
\definecolor{drawColor}{RGB}{0,0,0}

\node[text=drawColor,rotate= 90.00,anchor=base,inner sep=0pt, outer sep=0pt, scale=  1.10] at ( 13.08, 71.42) {Clustering Error};
\end{scope}
\begin{scope}
\path[clip] (  0.00,  0.00) rectangle (144.54,130.09);

\path[] ( 43.35, 69.99) rectangle (139.21,121.46);
\end{scope}
\begin{scope}
\path[clip] (  0.00,  0.00) rectangle (144.54,130.09);

\path[] ( 49.05, 90.13) rectangle ( 63.50,104.58);
\end{scope}
\begin{scope}
\path[clip] (  0.00,  0.00) rectangle (144.54,130.09);
\definecolor{drawColor}{RGB}{0,0,255}

\path[draw=drawColor,line width= 1.7pt,line join=round] ( 50.49, 97.36) -- ( 62.05, 97.36);
\end{scope}
\begin{scope}
\path[clip] (  0.00,  0.00) rectangle (144.54,130.09);
\definecolor{drawColor}{RGB}{0,0,255}

\path[draw=drawColor,line width= 1.7pt,line join=round] ( 50.49, 97.36) -- ( 62.05, 97.36);
\end{scope}
\begin{scope}
\path[clip] (  0.00,  0.00) rectangle (144.54,130.09);

\path[] ( 49.05, 75.68) rectangle ( 63.50, 90.13);
\end{scope}
\begin{scope}
\path[clip] (  0.00,  0.00) rectangle (144.54,130.09);
\definecolor{drawColor}{RGB}{0,0,0}

\path[draw=drawColor,line width= 1.7pt,line join=round] ( 50.49, 82.90) -- ( 62.05, 82.90);
\end{scope}
\begin{scope}
\path[clip] (  0.00,  0.00) rectangle (144.54,130.09);
\definecolor{drawColor}{RGB}{0,0,0}

\path[draw=drawColor,line width= 1.7pt,line join=round] ( 50.49, 82.90) -- ( 62.05, 82.90);
\end{scope}
\begin{scope}
\path[clip] (  0.00,  0.00) rectangle (144.54,130.09);
\definecolor{drawColor}{RGB}{0,0,0}

\node[text=drawColor,anchor=base west,inner sep=0pt, outer sep=0pt, scale=  0.93] at ( 65.31, 94.14) {{\sc Fsc} (this paper)};
\end{scope}
\begin{scope}
\path[clip] (  0.00,  0.00) rectangle (144.54,130.09);
\definecolor{drawColor}{RGB}{0,0,0}

\node[text=drawColor,anchor=base west,inner sep=0pt, outer sep=0pt, scale=  0.93] at ( 65.31, 79.68) {Best$(a$-$h)$};
\end{scope}
\end{tikzpicture}} \hspace{-.3cm}
\Scale[.6]{
\begin{tikzpicture}[x=1pt,y=1pt]
\definecolor{fillColor}{RGB}{255,255,255}
\path[use as bounding box,fill=fillColor,fill opacity=0.00] (0,0) rectangle (151.77,130.09);
\begin{scope}
\path[clip] (  0.00,  0.00) rectangle (151.77,130.09);
\definecolor{drawColor}{RGB}{255,255,255}
\definecolor{fillColor}{RGB}{255,255,255}

\path[draw=drawColor,line width= 0.6pt,line join=round,line cap=round,fill=fillColor] ( -0.00, -0.00) rectangle (151.77,130.09);
\end{scope}
\begin{scope}
\path[clip] ( 41.56, 30.92) rectangle (146.27,111.93);
\definecolor{fillColor}{gray}{0.92}

\path[fill=fillColor] ( 41.56, 30.92) rectangle (146.27,111.93);
\definecolor{drawColor}{RGB}{255,255,255}

\path[draw=drawColor,line width= 0.3pt,line join=round] ( 41.56, 32.93) --
	(146.27, 32.93);

\path[draw=drawColor,line width= 0.3pt,line join=round] ( 41.56, 49.67) --
	(146.27, 49.67);

\path[draw=drawColor,line width= 0.3pt,line join=round] ( 41.56, 66.40) --
	(146.27, 66.40);

\path[draw=drawColor,line width= 0.3pt,line join=round] ( 41.56, 83.14) --
	(146.27, 83.14);

\path[draw=drawColor,line width= 0.3pt,line join=round] ( 41.56, 99.88) --
	(146.27, 99.88);

\path[draw=drawColor,line width= 0.3pt,line join=round] ( 55.84, 30.92) --
	( 55.84,111.93);

\path[draw=drawColor,line width= 0.3pt,line join=round] ( 74.88, 30.92) --
	( 74.88,111.93);

\path[draw=drawColor,line width= 0.3pt,line join=round] ( 93.91, 30.92) --
	( 93.91,111.93);

\path[draw=drawColor,line width= 0.3pt,line join=round] (112.95, 30.92) --
	(112.95,111.93);

\path[draw=drawColor,line width= 0.3pt,line join=round] (131.99, 30.92) --
	(131.99,111.93);

\path[draw=drawColor,line width= 0.6pt,line join=round] ( 41.56, 41.30) --
	(146.27, 41.30);

\path[draw=drawColor,line width= 0.6pt,line join=round] ( 41.56, 58.04) --
	(146.27, 58.04);

\path[draw=drawColor,line width= 0.6pt,line join=round] ( 41.56, 74.77) --
	(146.27, 74.77);

\path[draw=drawColor,line width= 0.6pt,line join=round] ( 41.56, 91.51) --
	(146.27, 91.51);

\path[draw=drawColor,line width= 0.6pt,line join=round] ( 41.56,108.24) --
	(146.27,108.24);

\path[draw=drawColor,line width= 0.6pt,line join=round] ( 46.32, 30.92) --
	( 46.32,111.93);

\path[draw=drawColor,line width= 0.6pt,line join=round] ( 65.36, 30.92) --
	( 65.36,111.93);

\path[draw=drawColor,line width= 0.6pt,line join=round] ( 84.40, 30.92) --
	( 84.40,111.93);

\path[draw=drawColor,line width= 0.6pt,line join=round] (103.43, 30.92) --
	(103.43,111.93);

\path[draw=drawColor,line width= 0.6pt,line join=round] (122.47, 30.92) --
	(122.47,111.93);

\path[draw=drawColor,line width= 0.6pt,line join=round] (141.51, 30.92) --
	(141.51,111.93);
\definecolor{drawColor}{RGB}{0,0,0}

\path[draw=drawColor,line width= 1.7pt,line join=round] ( 46.32, 44.11) --
	( 65.36, 42.71) --
	( 84.40, 41.30) --
	(103.43, 41.30) --
	(122.47, 41.30) --
	(141.51, 41.30);
\definecolor{drawColor}{RGB}{0,0,255}

\path[draw=drawColor,line width= 1.7pt,line join=round] ( 46.32,101.75) --
	( 65.36, 45.65) --
	( 84.40, 44.85) --
	(103.43, 42.97) --
	(122.47, 41.97) --
	(141.51, 41.30);
\end{scope}
\begin{scope}
\path[clip] (  0.00,  0.00) rectangle (151.77,130.09);
\definecolor{drawColor}{RGB}{190,190,190}

\node[text=drawColor,anchor=base,inner sep=0pt, outer sep=0pt, scale=  0.79] at ( 27.59, 38.57) {0.000};

\node[text=drawColor,anchor=base,inner sep=0pt, outer sep=0pt, scale=  0.79] at ( 27.59, 55.31) {0.025};

\node[text=drawColor,anchor=base,inner sep=0pt, outer sep=0pt, scale=  0.79] at ( 27.59, 72.04) {0.050};

\node[text=drawColor,anchor=base,inner sep=0pt, outer sep=0pt, scale=  0.79] at ( 27.59, 88.78) {0.075};

\node[text=drawColor,anchor=base,inner sep=0pt, outer sep=0pt, scale=  0.79] at ( 27.59,105.52) {0.100};
\end{scope}
\begin{scope}
\path[clip] (  0.00,  0.00) rectangle (151.77,130.09);
\definecolor{drawColor}{gray}{0.20}

\path[draw=drawColor,line width= 0.6pt,line join=round] ( 38.81, 41.30) --
	( 41.56, 41.30);

\path[draw=drawColor,line width= 0.6pt,line join=round] ( 38.81, 58.04) --
	( 41.56, 58.04);

\path[draw=drawColor,line width= 0.6pt,line join=round] ( 38.81, 74.77) --
	( 41.56, 74.77);

\path[draw=drawColor,line width= 0.6pt,line join=round] ( 38.81, 91.51) --
	( 41.56, 91.51);

\path[draw=drawColor,line width= 0.6pt,line join=round] ( 38.81,108.24) --
	( 41.56,108.24);
\end{scope}
\begin{scope}
\path[clip] (  0.00,  0.00) rectangle (151.77,130.09);
\definecolor{drawColor}{gray}{0.20}

\path[draw=drawColor,line width= 0.6pt,line join=round] ( 46.32, 28.17) --
	( 46.32, 30.92);

\path[draw=drawColor,line width= 0.6pt,line join=round] ( 65.36, 28.17) --
	( 65.36, 30.92);

\path[draw=drawColor,line width= 0.6pt,line join=round] ( 84.40, 28.17) --
	( 84.40, 30.92);

\path[draw=drawColor,line width= 0.6pt,line join=round] (103.43, 28.17) --
	(103.43, 30.92);

\path[draw=drawColor,line width= 0.6pt,line join=round] (122.47, 28.17) --
	(122.47, 30.92);

\path[draw=drawColor,line width= 0.6pt,line join=round] (141.51, 28.17) --
	(141.51, 30.92);
\end{scope}
\begin{scope}
\path[clip] (  0.00,  0.00) rectangle (151.77,130.09);
\definecolor{drawColor}{RGB}{190,190,190}

\node[text=drawColor,anchor=base,inner sep=0pt, outer sep=0pt, scale=  0.79] at ( 46.32, 20.52) {6};

\node[text=drawColor,anchor=base,inner sep=0pt, outer sep=0pt, scale=  0.79] at ( 65.36, 20.52) {10};

\node[text=drawColor,anchor=base,inner sep=0pt, outer sep=0pt, scale=  0.79] at ( 84.40, 20.52) {15};

\node[text=drawColor,anchor=base,inner sep=0pt, outer sep=0pt, scale=  0.79] at (103.43, 20.52) {20};

\node[text=drawColor,anchor=base,inner sep=0pt, outer sep=0pt, scale=  0.79] at (122.47, 20.52) {25};

\node[text=drawColor,anchor=base,inner sep=0pt, outer sep=0pt, scale=  0.79] at (141.51, 20.52) {30};
\end{scope}
\begin{scope}
\path[clip] (  0.00,  0.00) rectangle (151.77,130.09);
\definecolor{drawColor}{RGB}{0,0,0}

\node[text=drawColor,anchor=base,inner sep=0pt, outer sep=0pt, scale=  1.10] at ( 93.91,  7.44) {Columns per Subspace};
\end{scope}
\end{tikzpicture}} \hspace{-.3cm}
\Scale[.6]{
\begin{tikzpicture}[x=1pt,y=1pt]
\definecolor{fillColor}{RGB}{255,255,255}
\path[use as bounding box,fill=fillColor,fill opacity=0.00] (0,0) rectangle (144.54,133.70);
\begin{scope}
\path[clip] (  0.00,  0.00) rectangle (144.54,133.70);
\definecolor{drawColor}{RGB}{255,255,255}
\definecolor{fillColor}{RGB}{255,255,255}

\path[draw=drawColor,line width= 0.6pt,line join=round,line cap=round,fill=fillColor] (  0.00,  0.00) rectangle (144.54,133.70);
\end{scope}
\begin{scope}
\path[clip] ( 33.64, 30.92) rectangle (139.04,115.54);
\definecolor{fillColor}{gray}{0.92}

\path[fill=fillColor] ( 33.64, 30.92) rectangle (139.04,115.54);
\definecolor{drawColor}{RGB}{255,255,255}

\path[draw=drawColor,line width= 0.3pt,line join=round] ( 33.64, 44.63) --
	(139.04, 44.63);

\path[draw=drawColor,line width= 0.3pt,line join=round] ( 33.64, 64.36) --
	(139.04, 64.36);

\path[draw=drawColor,line width= 0.3pt,line join=round] ( 33.64, 84.08) --
	(139.04, 84.08);

\path[draw=drawColor,line width= 0.3pt,line join=round] ( 33.64,103.80) --
	(139.04,103.80);

\path[draw=drawColor,line width= 0.3pt,line join=round] ( 46.69, 30.92) --
	( 46.69,115.54);

\path[draw=drawColor,line width= 0.3pt,line join=round] ( 66.52, 30.92) --
	( 66.52,115.54);

\path[draw=drawColor,line width= 0.3pt,line join=round] ( 92.95, 30.92) --
	( 92.95,115.54);

\path[draw=drawColor,line width= 0.3pt,line join=round] (119.38, 30.92) --
	(119.38,115.54);

\path[draw=drawColor,line width= 0.6pt,line join=round] ( 33.64, 34.77) --
	(139.04, 34.77);

\path[draw=drawColor,line width= 0.6pt,line join=round] ( 33.64, 54.49) --
	(139.04, 54.49);

\path[draw=drawColor,line width= 0.6pt,line join=round] ( 33.64, 74.22) --
	(139.04, 74.22);

\path[draw=drawColor,line width= 0.6pt,line join=round] ( 33.64, 93.94) --
	(139.04, 93.94);

\path[draw=drawColor,line width= 0.6pt,line join=round] ( 33.64,113.66) --
	(139.04,113.66);

\path[draw=drawColor,line width= 0.6pt,line join=round] ( 40.09, 30.92) --
	( 40.09,115.54);

\path[draw=drawColor,line width= 0.6pt,line join=round] ( 53.30, 30.92) --
	( 53.30,115.54);

\path[draw=drawColor,line width= 0.6pt,line join=round] ( 79.73, 30.92) --
	( 79.73,115.54);

\path[draw=drawColor,line width= 0.6pt,line join=round] (106.17, 30.92) --
	(106.17,115.54);

\path[draw=drawColor,line width= 0.6pt,line join=round] (132.60, 30.92) --
	(132.60,115.54);
\definecolor{drawColor}{RGB}{0,0,0}

\path[draw=drawColor,line width= 1.7pt,line join=round] ( 40.09,111.69) --
	( 53.30,105.04) --
	( 66.52, 95.17) --
	( 79.73,103.80) --
	( 92.95, 66.82) --
	(106.17, 34.77) --
	(119.38, 34.77) --
	(132.60, 34.77);
\definecolor{drawColor}{RGB}{0,0,255}

\path[draw=drawColor,line width= 1.7pt,line join=round] ( 40.09,107.75) --
	( 53.30, 42.17) --
	( 66.52, 42.17) --
	( 79.73, 47.23) --
	( 92.95, 47.59) --
	(106.17, 34.77) --
	(119.38, 34.77) --
	(132.60, 34.77);
\definecolor{drawColor}{RGB}{0,0,0}

\node[text=drawColor,anchor=base,inner sep=0pt, outer sep=0pt, scale=  1.10] at (112.77, 95.07) {${\rm n}_{\rm k}=20$};
\end{scope}
\begin{scope}
\path[clip] (  0.00,  0.00) rectangle (144.54,133.70);
\definecolor{drawColor}{RGB}{190,190,190}

\node[text=drawColor,anchor=base,inner sep=0pt, outer sep=0pt, scale=  0.79] at ( 23.63, 32.04) {0.0};

\node[text=drawColor,anchor=base,inner sep=0pt, outer sep=0pt, scale=  0.79] at ( 23.63, 51.77) {0.2};

\node[text=drawColor,anchor=base,inner sep=0pt, outer sep=0pt, scale=  0.79] at ( 23.63, 71.49) {0.4};

\node[text=drawColor,anchor=base,inner sep=0pt, outer sep=0pt, scale=  0.79] at ( 23.63, 91.21) {0.6};

\node[text=drawColor,anchor=base,inner sep=0pt, outer sep=0pt, scale=  0.79] at ( 23.63,110.94) {0.8};
\end{scope}
\begin{scope}
\path[clip] (  0.00,  0.00) rectangle (144.54,133.70);
\definecolor{drawColor}{gray}{0.20}

\path[draw=drawColor,line width= 0.6pt,line join=round] ( 30.89, 34.77) --
	( 33.64, 34.77);

\path[draw=drawColor,line width= 0.6pt,line join=round] ( 30.89, 54.49) --
	( 33.64, 54.49);

\path[draw=drawColor,line width= 0.6pt,line join=round] ( 30.89, 74.22) --
	( 33.64, 74.22);

\path[draw=drawColor,line width= 0.6pt,line join=round] ( 30.89, 93.94) --
	( 33.64, 93.94);

\path[draw=drawColor,line width= 0.6pt,line join=round] ( 30.89,113.66) --
	( 33.64,113.66);
\end{scope}
\begin{scope}
\path[clip] (  0.00,  0.00) rectangle (144.54,133.70);
\definecolor{drawColor}{gray}{0.20}

\path[draw=drawColor,line width= 0.6pt,line join=round] ( 40.09, 28.17) --
	( 40.09, 30.92);

\path[draw=drawColor,line width= 0.6pt,line join=round] ( 53.30, 28.17) --
	( 53.30, 30.92);

\path[draw=drawColor,line width= 0.6pt,line join=round] ( 79.73, 28.17) --
	( 79.73, 30.92);

\path[draw=drawColor,line width= 0.6pt,line join=round] (106.17, 28.17) --
	(106.17, 30.92);

\path[draw=drawColor,line width= 0.6pt,line join=round] (132.60, 28.17) --
	(132.60, 30.92);
\end{scope}
\begin{scope}
\path[clip] (  0.00,  0.00) rectangle (144.54,133.70);
\definecolor{drawColor}{RGB}{190,190,190}

\node[text=drawColor,anchor=base,inner sep=0pt, outer sep=0pt, scale=  0.79] at ( 40.09, 20.52) {.06};

\node[text=drawColor,anchor=base,inner sep=0pt, outer sep=0pt, scale=  0.79] at ( 53.30, 20.52) {.1};

\node[text=drawColor,anchor=base,inner sep=0pt, outer sep=0pt, scale=  0.79] at ( 79.73, 20.52) {.3};

\node[text=drawColor,anchor=base,inner sep=0pt, outer sep=0pt, scale=  0.79] at (106.17, 20.52) {.5};

\node[text=drawColor,anchor=base,inner sep=0pt, outer sep=0pt, scale=  0.79] at (132.60, 20.52) {.7};
\end{scope}
\begin{scope}
\path[clip] (  0.00,  0.00) rectangle (144.54,133.70);
\definecolor{drawColor}{RGB}{0,0,0}

\node[text=drawColor,anchor=base,inner sep=0pt, outer sep=0pt, scale=  1.10] at ( 86.34,  7.44) {Sampling Rate};
\end{scope}
\end{tikzpicture}} \hspace{-.3cm}
\Scale[.6]{
\begin{tikzpicture}[x=1pt,y=1pt]
\definecolor{fillColor}{RGB}{255,255,255}
\path[use as bounding box,fill=fillColor,fill opacity=0.00] (0,0) rectangle (144.54,133.70);
\begin{scope}
\path[clip] (  0.00,  0.00) rectangle (144.54,133.70);
\definecolor{drawColor}{RGB}{255,255,255}
\definecolor{fillColor}{RGB}{255,255,255}

\path[draw=drawColor,line width= 0.6pt,line join=round,line cap=round,fill=fillColor] (  0.00,  0.00) rectangle (144.54,133.70);
\end{scope}
\begin{scope}
\path[clip] ( 33.64, 30.92) rectangle (139.04,115.54);
\definecolor{fillColor}{gray}{0.92}

\path[fill=fillColor] ( 33.64, 30.92) rectangle (139.04,115.54);
\definecolor{drawColor}{RGB}{255,255,255}

\path[draw=drawColor,line width= 0.3pt,line join=round] ( 33.64, 44.89) --
	(139.04, 44.89);

\path[draw=drawColor,line width= 0.3pt,line join=round] ( 33.64, 65.13) --
	(139.04, 65.13);

\path[draw=drawColor,line width= 0.3pt,line join=round] ( 33.64, 85.38) --
	(139.04, 85.38);

\path[draw=drawColor,line width= 0.3pt,line join=round] ( 33.64,105.62) --
	(139.04,105.62);

\path[draw=drawColor,line width= 0.3pt,line join=round] ( 46.69, 30.92) --
	( 46.69,115.54);

\path[draw=drawColor,line width= 0.3pt,line join=round] ( 66.52, 30.92) --
	( 66.52,115.54);

\path[draw=drawColor,line width= 0.3pt,line join=round] ( 92.95, 30.92) --
	( 92.95,115.54);

\path[draw=drawColor,line width= 0.3pt,line join=round] (119.38, 30.92) --
	(119.38,115.54);

\path[draw=drawColor,line width= 0.6pt,line join=round] ( 33.64, 34.77) --
	(139.04, 34.77);

\path[draw=drawColor,line width= 0.6pt,line join=round] ( 33.64, 55.01) --
	(139.04, 55.01);

\path[draw=drawColor,line width= 0.6pt,line join=round] ( 33.64, 75.26) --
	(139.04, 75.26);

\path[draw=drawColor,line width= 0.6pt,line join=round] ( 33.64, 95.50) --
	(139.04, 95.50);

\path[draw=drawColor,line width= 0.6pt,line join=round] ( 40.09, 30.92) --
	( 40.09,115.54);

\path[draw=drawColor,line width= 0.6pt,line join=round] ( 53.30, 30.92) --
	( 53.30,115.54);

\path[draw=drawColor,line width= 0.6pt,line join=round] ( 79.73, 30.92) --
	( 79.73,115.54);

\path[draw=drawColor,line width= 0.6pt,line join=round] (106.17, 30.92) --
	(106.17,115.54);

\path[draw=drawColor,line width= 0.6pt,line join=round] (132.60, 30.92) --
	(132.60,115.54);
\definecolor{drawColor}{RGB}{0,0,0}

\path[draw=drawColor,line width= 1.7pt,line join=round] ( 40.09,109.67) --
	( 53.30,102.08) --
	( 66.52,103.09) --
	( 79.73, 34.77) --
	( 92.95, 34.77) --
	(106.17, 34.77) --
	(119.38, 34.77) --
	(132.60, 34.77);
\definecolor{drawColor}{RGB}{0,0,255}

\path[draw=drawColor,line width= 1.7pt,line join=round] ( 40.09,111.69) --
	( 53.30, 51.47) --
	( 66.52, 52.08) --
	( 79.73, 34.77) --
	( 92.95, 34.77) --
	(106.17, 34.77) --
	(119.38, 34.77) --
	(132.60, 34.77);
\definecolor{drawColor}{RGB}{0,0,0}

\node[text=drawColor,anchor=base,inner sep=0pt, outer sep=0pt, scale=  1.10] at (106.17, 96.76) {${\rm n}_{\rm k}=50$};
\end{scope}
\begin{scope}
\path[clip] (  0.00,  0.00) rectangle (144.54,133.70);
\definecolor{drawColor}{RGB}{190,190,190}

\node[text=drawColor,anchor=base,inner sep=0pt, outer sep=0pt, scale=  0.79] at ( 23.63, 32.04) {0.0};

\node[text=drawColor,anchor=base,inner sep=0pt, outer sep=0pt, scale=  0.79] at ( 23.63, 52.29) {0.2};

\node[text=drawColor,anchor=base,inner sep=0pt, outer sep=0pt, scale=  0.79] at ( 23.63, 72.53) {0.4};

\node[text=drawColor,anchor=base,inner sep=0pt, outer sep=0pt, scale=  0.79] at ( 23.63, 92.77) {0.6};
\end{scope}
\begin{scope}
\path[clip] (  0.00,  0.00) rectangle (144.54,133.70);
\definecolor{drawColor}{gray}{0.20}

\path[draw=drawColor,line width= 0.6pt,line join=round] ( 30.89, 34.77) --
	( 33.64, 34.77);

\path[draw=drawColor,line width= 0.6pt,line join=round] ( 30.89, 55.01) --
	( 33.64, 55.01);

\path[draw=drawColor,line width= 0.6pt,line join=round] ( 30.89, 75.26) --
	( 33.64, 75.26);

\path[draw=drawColor,line width= 0.6pt,line join=round] ( 30.89, 95.50) --
	( 33.64, 95.50);
\end{scope}
\begin{scope}
\path[clip] (  0.00,  0.00) rectangle (144.54,133.70);
\definecolor{drawColor}{gray}{0.20}

\path[draw=drawColor,line width= 0.6pt,line join=round] ( 40.09, 28.17) --
	( 40.09, 30.92);

\path[draw=drawColor,line width= 0.6pt,line join=round] ( 53.30, 28.17) --
	( 53.30, 30.92);

\path[draw=drawColor,line width= 0.6pt,line join=round] ( 79.73, 28.17) --
	( 79.73, 30.92);

\path[draw=drawColor,line width= 0.6pt,line join=round] (106.17, 28.17) --
	(106.17, 30.92);

\path[draw=drawColor,line width= 0.6pt,line join=round] (132.60, 28.17) --
	(132.60, 30.92);
\end{scope}
\begin{scope}
\path[clip] (  0.00,  0.00) rectangle (144.54,133.70);
\definecolor{drawColor}{RGB}{190,190,190}

\node[text=drawColor,anchor=base,inner sep=0pt, outer sep=0pt, scale=  0.79] at ( 40.09, 20.52) {.06};

\node[text=drawColor,anchor=base,inner sep=0pt, outer sep=0pt, scale=  0.79] at ( 53.30, 20.52) {.1};

\node[text=drawColor,anchor=base,inner sep=0pt, outer sep=0pt, scale=  0.79] at ( 79.73, 20.52) {.3};

\node[text=drawColor,anchor=base,inner sep=0pt, outer sep=0pt, scale=  0.79] at (106.17, 20.52) {.5};

\node[text=drawColor,anchor=base,inner sep=0pt, outer sep=0pt, scale=  0.79] at (132.60, 20.52) {.7};
\end{scope}
\begin{scope}
\path[clip] (  0.00,  0.00) rectangle (144.54,133.70);
\definecolor{drawColor}{RGB}{0,0,0}

\node[text=drawColor,anchor=base,inner sep=0pt, outer sep=0pt, scale=  1.10] at ( 86.34,  7.44) {Sampling Rate};
\end{scope}
\end{tikzpicture}}
\caption{Clustering error of \FSC\ and the best algorithm among $(a)$-$(h)$ in each trial. Notice the different scales. With full-data \FSC\ is rarely outperformed by other algorithms, and only by a marginal amount. In contrast, when data is missing, \FSC\ outperforms other algorithms by a wide margin. For example, with $\n_\k=20$ and $\p=0.1$, \FSC\ achieves $7.5\%$ error, while the next best algorithm achieves $71.25\%$.}
\label{results2Fig}
\end{figure}
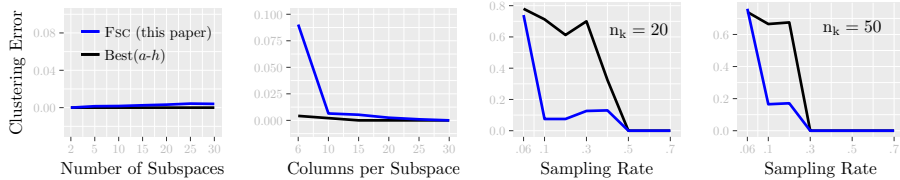

{\bf Effect of the number of subspaces and data points.} Figure \ref{results2Fig} shows that like the state-of-the-art, \FSC\ is very robust to the number of subspaces $\K$. Recall that in our default settings, $\r=5$, so $\K \geq 20$ produces a full-rank data matrix $\X$. Figure \ref{results2Fig} also evaluates the performance of \FSC\ as a function of the columns per subspace $\n_\k$. Since $\r=5$, $\n_\k=6$ is information-theoretically necessary for subspace clustering. \FSC\ only requires little more than that to perform as well as the state-of-the-art.

{\bf Effect of missing data.} There is a tradeoff between the number of columns per subspace $\n_\k$ and the samples per column $\p$ required for subspace clustering \cite{infoTheoretic}. The larger $\n_\k$, the lower $\p$ may be, and vice versa. Figure \ref{results2Fig} evaluates the performance of \FSC\ as a function of $\p$ with $\n_\k=20,50$ (few and many columns). Notice that if the sampling rate $\p$ is high (few missing data), then \FSC\ performs as well as the state-of-the-art, and much better if the sampling rate $\p$ is low (many missing data); see for example $\n_\k=20$ and $\p=0.1$, where the best among algorithms $(a)$-$(h)$ gets $71.25\%$ error, which is as good as random guessing (because there are $\K=4$ subspaces in our default settings). In contrast, \FSC\ gets $7.5\%$ error. Notice that $\p=0.1$ is very close to the exact information-theoretic minimum sampling rate $\p=(\r+1)/\d=0.06$ \cite{infoTheoretic}. Similar to noise, if there is much missing data the first term in \eqref{fscEq} and \eqref{ifscEq} will carry less weight, which we can compensate by making $\lambdaa$ smaller.


\subsection{Real Data Experiments}
\begin{figure}
\centering
\includegraphics[width=5cm]{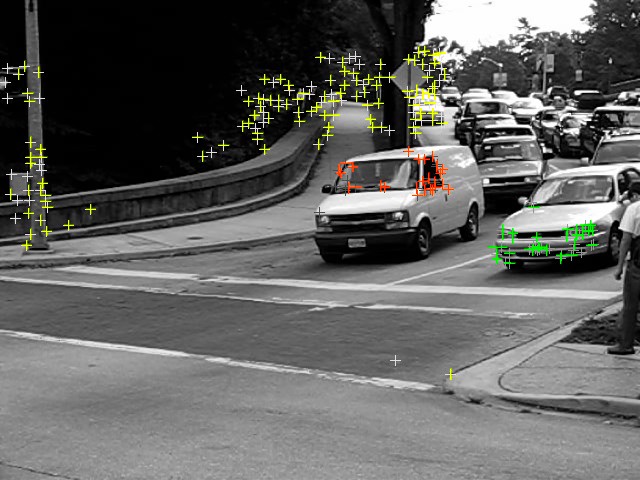}
\caption{Points tracked in a frame from Hopkins 155 \cite{hopkins}.}
\label{hopkinsEgFig}
\end{figure}

{\bf Motion Segmentation.}
It is well-known that the locations over time of a rigidly moving object approximately lie in a $3$-dimensional affine subspace \cite{kanade,kanatani} (which can be thought as a $4$-dimensional subspace whose fourth component accounts for the offset). Hence, by tracking points in a video, and subspace clustering them, we can segment the multiple moving objects appearing in the video. In this experiment we test \FSC\ on this task, using the Hopkins 155 dataset \cite{hopkins}, containing sequences of points tracked over time in $155$ videos; Figure \ref{hopkinsEgFig} shows a sample frame. Each video contains either $\K=2$ or $\K=3$ objects. On average, each object is tracked on $\n_\k=133$ points (described by two coordinates) over $29$ frames, producing vectors in ambient dimension $\d=58$.

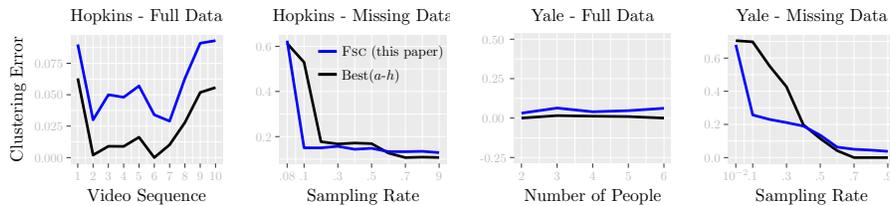
\begin{figure}[b]
\centering
\Scale[.6]{
\begin{tikzpicture}[x=1pt,y=1pt]
\definecolor{fillColor}{RGB}{255,255,255}
\path[use as bounding box,fill=fillColor,fill opacity=0.00] (0,0) rectangle (144.54,133.70);
\begin{scope}
\path[clip] (  0.00,  0.00) rectangle (144.54,133.70);
\definecolor{drawColor}{RGB}{255,255,255}
\definecolor{fillColor}{RGB}{255,255,255}

\path[draw=drawColor,line width= 0.6pt,line join=round,line cap=round,fill=fillColor] ( -0.00,  0.00) rectangle (144.54,133.70);
\end{scope}
\begin{scope}
\path[clip] ( 43.51, 30.92) rectangle (139.04,112.08);
\definecolor{fillColor}{gray}{0.92}

\path[fill=fillColor] ( 43.51, 30.92) rectangle (139.04,112.08);
\definecolor{drawColor}{RGB}{255,255,255}

\path[draw=drawColor,line width= 0.3pt,line join=round] ( 43.51, 44.53) --
	(139.04, 44.53);

\path[draw=drawColor,line width= 0.3pt,line join=round] ( 43.51, 64.36) --
	(139.04, 64.36);

\path[draw=drawColor,line width= 0.3pt,line join=round] ( 43.51, 84.19) --
	(139.04, 84.19);

\path[draw=drawColor,line width= 0.3pt,line join=round] ( 43.51,104.03) --
	(139.04,104.03);

\path[draw=drawColor,line width= 0.3pt,line join=round] ( 52.67, 30.92) --
	( 52.67,112.08);

\path[draw=drawColor,line width= 0.3pt,line join=round] ( 62.32, 30.92) --
	( 62.32,112.08);

\path[draw=drawColor,line width= 0.3pt,line join=round] ( 71.97, 30.92) --
	( 71.97,112.08);

\path[draw=drawColor,line width= 0.3pt,line join=round] ( 81.62, 30.92) --
	( 81.62,112.08);

\path[draw=drawColor,line width= 0.3pt,line join=round] ( 91.27, 30.92) --
	( 91.27,112.08);

\path[draw=drawColor,line width= 0.3pt,line join=round] (100.92, 30.92) --
	(100.92,112.08);

\path[draw=drawColor,line width= 0.3pt,line join=round] (110.57, 30.92) --
	(110.57,112.08);

\path[draw=drawColor,line width= 0.3pt,line join=round] (120.22, 30.92) --
	(120.22,112.08);

\path[draw=drawColor,line width= 0.3pt,line join=round] (129.87, 30.92) --
	(129.87,112.08);

\path[draw=drawColor,line width= 0.6pt,line join=round] ( 43.51, 34.61) --
	(139.04, 34.61);

\path[draw=drawColor,line width= 0.6pt,line join=round] ( 43.51, 54.45) --
	(139.04, 54.45);

\path[draw=drawColor,line width= 0.6pt,line join=round] ( 43.51, 74.28) --
	(139.04, 74.28);

\path[draw=drawColor,line width= 0.6pt,line join=round] ( 43.51, 94.11) --
	(139.04, 94.11);

\path[draw=drawColor,line width= 0.6pt,line join=round] ( 47.85, 30.92) --
	( 47.85,112.08);

\path[draw=drawColor,line width= 0.6pt,line join=round] ( 57.50, 30.92) --
	( 57.50,112.08);

\path[draw=drawColor,line width= 0.6pt,line join=round] ( 67.15, 30.92) --
	( 67.15,112.08);

\path[draw=drawColor,line width= 0.6pt,line join=round] ( 76.80, 30.92) --
	( 76.80,112.08);

\path[draw=drawColor,line width= 0.6pt,line join=round] ( 86.45, 30.92) --
	( 86.45,112.08);

\path[draw=drawColor,line width= 0.6pt,line join=round] ( 96.10, 30.92) --
	( 96.10,112.08);

\path[draw=drawColor,line width= 0.6pt,line join=round] (105.75, 30.92) --
	(105.75,112.08);

\path[draw=drawColor,line width= 0.6pt,line join=round] (115.40, 30.92) --
	(115.40,112.08);

\path[draw=drawColor,line width= 0.6pt,line join=round] (125.05, 30.92) --
	(125.05,112.08);

\path[draw=drawColor,line width= 0.6pt,line join=round] (134.70, 30.92) --
	(134.70,112.08);
\definecolor{drawColor}{RGB}{0,0,0}

\path[draw=drawColor,line width= 1.7pt,line join=round] ( 47.85, 84.59) --
	( 57.50, 36.20) --
	( 67.15, 41.75) --
	( 76.80, 41.67) --
	( 86.45, 47.47) --
	( 96.10, 34.61) --
	(105.75, 42.71) --
	(115.40, 56.75) --
	(125.05, 75.71) --
	(134.70, 78.80);
\definecolor{drawColor}{RGB}{0,0,255}

\path[draw=drawColor,line width= 1.7pt,line join=round] ( 47.85,106.01) --
	( 57.50, 58.41) --
	( 67.15, 74.28) --
	( 76.80, 72.69) --
	( 86.45, 79.83) --
	( 96.10, 61.59) --
	(105.75, 57.62) --
	(115.40, 84.59) --
	(125.05,106.80) --
	(134.70,108.39);
\end{scope}
\begin{scope}
\path[clip] (  0.00,  0.00) rectangle (144.54,133.70);
\definecolor{drawColor}{RGB}{190,190,190}

\node[text=drawColor,anchor=base,inner sep=0pt, outer sep=0pt, scale=  0.79] at ( 29.54, 31.89) {0.000};

\node[text=drawColor,anchor=base,inner sep=0pt, outer sep=0pt, scale=  0.79] at ( 29.54, 51.72) {0.025};

\node[text=drawColor,anchor=base,inner sep=0pt, outer sep=0pt, scale=  0.79] at ( 29.54, 71.55) {0.050};

\node[text=drawColor,anchor=base,inner sep=0pt, outer sep=0pt, scale=  0.79] at ( 29.54, 91.38) {0.075};
\end{scope}
\begin{scope}
\path[clip] (  0.00,  0.00) rectangle (144.54,133.70);
\definecolor{drawColor}{gray}{0.20}

\path[draw=drawColor,line width= 0.6pt,line join=round] ( 40.76, 34.61) --
	( 43.51, 34.61);

\path[draw=drawColor,line width= 0.6pt,line join=round] ( 40.76, 54.45) --
	( 43.51, 54.45);

\path[draw=drawColor,line width= 0.6pt,line join=round] ( 40.76, 74.28) --
	( 43.51, 74.28);

\path[draw=drawColor,line width= 0.6pt,line join=round] ( 40.76, 94.11) --
	( 43.51, 94.11);
\end{scope}
\begin{scope}
\path[clip] (  0.00,  0.00) rectangle (144.54,133.70);
\definecolor{drawColor}{gray}{0.20}

\path[draw=drawColor,line width= 0.6pt,line join=round] ( 47.85, 28.17) --
	( 47.85, 30.92);

\path[draw=drawColor,line width= 0.6pt,line join=round] ( 57.50, 28.17) --
	( 57.50, 30.92);

\path[draw=drawColor,line width= 0.6pt,line join=round] ( 67.15, 28.17) --
	( 67.15, 30.92);

\path[draw=drawColor,line width= 0.6pt,line join=round] ( 76.80, 28.17) --
	( 76.80, 30.92);

\path[draw=drawColor,line width= 0.6pt,line join=round] ( 86.45, 28.17) --
	( 86.45, 30.92);

\path[draw=drawColor,line width= 0.6pt,line join=round] ( 96.10, 28.17) --
	( 96.10, 30.92);

\path[draw=drawColor,line width= 0.6pt,line join=round] (105.75, 28.17) --
	(105.75, 30.92);

\path[draw=drawColor,line width= 0.6pt,line join=round] (115.40, 28.17) --
	(115.40, 30.92);

\path[draw=drawColor,line width= 0.6pt,line join=round] (125.05, 28.17) --
	(125.05, 30.92);

\path[draw=drawColor,line width= 0.6pt,line join=round] (134.70, 28.17) --
	(134.70, 30.92);
\end{scope}
\begin{scope}
\path[clip] (  0.00,  0.00) rectangle (144.54,133.70);
\definecolor{drawColor}{RGB}{190,190,190}

\node[text=drawColor,anchor=base,inner sep=0pt, outer sep=0pt, scale=  0.79] at ( 47.85, 20.52) {1};

\node[text=drawColor,anchor=base,inner sep=0pt, outer sep=0pt, scale=  0.79] at ( 57.50, 20.52) {2};

\node[text=drawColor,anchor=base,inner sep=0pt, outer sep=0pt, scale=  0.79] at ( 67.15, 20.52) {3};

\node[text=drawColor,anchor=base,inner sep=0pt, outer sep=0pt, scale=  0.79] at ( 76.80, 20.52) {4};

\node[text=drawColor,anchor=base,inner sep=0pt, outer sep=0pt, scale=  0.79] at ( 86.45, 20.52) {5};

\node[text=drawColor,anchor=base,inner sep=0pt, outer sep=0pt, scale=  0.79] at ( 96.10, 20.52) {6};

\node[text=drawColor,anchor=base,inner sep=0pt, outer sep=0pt, scale=  0.79] at (105.75, 20.52) {7};

\node[text=drawColor,anchor=base,inner sep=0pt, outer sep=0pt, scale=  0.79] at (115.40, 20.52) {8};

\node[text=drawColor,anchor=base,inner sep=0pt, outer sep=0pt, scale=  0.79] at (125.05, 20.52) {9};

\node[text=drawColor,anchor=base,inner sep=0pt, outer sep=0pt, scale=  0.79] at (134.70, 20.52) {10};
\end{scope}
\begin{scope}
\path[clip] (  0.00,  0.00) rectangle (144.54,133.70);
\definecolor{drawColor}{RGB}{0,0,0}

\node[text=drawColor,anchor=base,inner sep=0pt, outer sep=0pt, scale=  1.10] at ( 91.27,  7.44) {Video Sequence};
\end{scope}
\begin{scope}
\path[clip] (  0.00,  0.00) rectangle (144.54,133.70);
\definecolor{drawColor}{RGB}{0,0,0}

\node[text=drawColor,rotate= 90.00,anchor=base,inner sep=0pt, outer sep=0pt, scale=  1.10] at ( 13.08, 71.50) {Clustering Error};
\end{scope}
\begin{scope}
\path[clip] (  0.00,  0.00) rectangle (144.54,133.70);
\definecolor{drawColor}{RGB}{0,0,0}

\node[text=drawColor,anchor=base,inner sep=0pt, outer sep=0pt, scale=  1.10] at ( 91.27,120.62) {Hopkins - Full Data};
\end{scope}
\end{tikzpicture}} \hspace{-.3cm}
\Scale[.6]{
\begin{tikzpicture}[x=1pt,y=1pt]
\definecolor{fillColor}{RGB}{255,255,255}
\path[use as bounding box,fill=fillColor,fill opacity=0.00] (0,0) rectangle (144.54,133.70);
\begin{scope}
\path[clip] (  0.00,  0.00) rectangle (144.54,133.70);
\definecolor{drawColor}{RGB}{255,255,255}
\definecolor{fillColor}{RGB}{255,255,255}

\path[draw=drawColor,line width= 0.6pt,line join=round,line cap=round,fill=fillColor] (  0.00,  0.00) rectangle (144.54,133.70);
\end{scope}
\begin{scope}
\path[clip] ( 33.64, 30.92) rectangle (139.04,112.08);
\definecolor{fillColor}{gray}{0.92}

\path[fill=fillColor] ( 33.64, 30.92) rectangle (139.04,112.08);
\definecolor{drawColor}{RGB}{255,255,255}

\path[draw=drawColor,line width= 0.3pt,line join=round] ( 33.64, 33.62) --
	(139.04, 33.62);

\path[draw=drawColor,line width= 0.3pt,line join=round] ( 33.64, 62.10) --
	(139.04, 62.10);

\path[draw=drawColor,line width= 0.3pt,line join=round] ( 33.64, 90.59) --
	(139.04, 90.59);

\path[draw=drawColor,line width= 0.3pt,line join=round] ( 43.76, 30.92) --
	( 43.76,112.08);

\path[draw=drawColor,line width= 0.3pt,line join=round] ( 59.73, 30.92) --
	( 59.73,112.08);

\path[draw=drawColor,line width= 0.3pt,line join=round] ( 81.02, 30.92) --
	( 81.02,112.08);

\path[draw=drawColor,line width= 0.3pt,line join=round] (102.31, 30.92) --
	(102.31,112.08);

\path[draw=drawColor,line width= 0.3pt,line join=round] (123.60, 30.92) --
	(123.60,112.08);

\path[draw=drawColor,line width= 0.6pt,line join=round] ( 33.64, 47.86) --
	(139.04, 47.86);

\path[draw=drawColor,line width= 0.6pt,line join=round] ( 33.64, 76.34) --
	(139.04, 76.34);

\path[draw=drawColor,line width= 0.6pt,line join=round] ( 33.64,104.83) --
	(139.04,104.83);

\path[draw=drawColor,line width= 0.6pt,line join=round] ( 38.43, 30.92) --
	( 38.43,112.08);

\path[draw=drawColor,line width= 0.6pt,line join=round] ( 49.08, 30.92) --
	( 49.08,112.08);

\path[draw=drawColor,line width= 0.6pt,line join=round] ( 70.37, 30.92) --
	( 70.37,112.08);

\path[draw=drawColor,line width= 0.6pt,line join=round] ( 91.66, 30.92) --
	( 91.66,112.08);

\path[draw=drawColor,line width= 0.6pt,line join=round] (112.96, 30.92) --
	(112.96,112.08);

\path[draw=drawColor,line width= 0.6pt,line join=round] (134.25, 30.92) --
	(134.25,112.08);
\definecolor{drawColor}{RGB}{0,0,0}

\path[draw=drawColor,line width= 1.7pt,line join=round] ( 38.43,106.61) --
	( 49.08, 94.77) --
	( 59.73, 44.64) --
	( 70.37, 43.20) --
	( 81.02, 43.79) --
	( 91.66, 43.42) --
	(102.31, 37.46) --
	(112.96, 34.61) --
	(123.60, 34.90) --
	(134.25, 34.67);
\definecolor{drawColor}{RGB}{0,0,255}

\path[draw=drawColor,line width= 1.7pt,line join=round] ( 38.43,108.39) --
	( 49.08, 40.78) --
	( 59.73, 40.74) --
	( 70.37, 41.72) --
	( 81.02, 39.85) --
	( 91.66, 40.47) --
	(102.31, 38.35) --
	(112.96, 38.33) --
	(123.60, 38.56) --
	(134.25, 37.74);
\end{scope}
\begin{scope}
\path[clip] (  0.00,  0.00) rectangle (144.54,133.70);
\definecolor{drawColor}{RGB}{190,190,190}

\node[text=drawColor,anchor=base,inner sep=0pt, outer sep=0pt, scale=  0.79] at ( 23.63, 45.13) {0.2};

\node[text=drawColor,anchor=base,inner sep=0pt, outer sep=0pt, scale=  0.79] at ( 23.63, 73.62) {0.4};

\node[text=drawColor,anchor=base,inner sep=0pt, outer sep=0pt, scale=  0.79] at ( 23.63,102.10) {0.6};
\end{scope}
\begin{scope}
\path[clip] (  0.00,  0.00) rectangle (144.54,133.70);
\definecolor{drawColor}{gray}{0.20}

\path[draw=drawColor,line width= 0.6pt,line join=round] ( 30.89, 47.86) --
	( 33.64, 47.86);

\path[draw=drawColor,line width= 0.6pt,line join=round] ( 30.89, 76.34) --
	( 33.64, 76.34);

\path[draw=drawColor,line width= 0.6pt,line join=round] ( 30.89,104.83) --
	( 33.64,104.83);
\end{scope}
\begin{scope}
\path[clip] (  0.00,  0.00) rectangle (144.54,133.70);
\definecolor{drawColor}{gray}{0.20}

\path[draw=drawColor,line width= 0.6pt,line join=round] ( 38.43, 28.17) --
	( 38.43, 30.92);

\path[draw=drawColor,line width= 0.6pt,line join=round] ( 49.08, 28.17) --
	( 49.08, 30.92);

\path[draw=drawColor,line width= 0.6pt,line join=round] ( 70.37, 28.17) --
	( 70.37, 30.92);

\path[draw=drawColor,line width= 0.6pt,line join=round] ( 91.66, 28.17) --
	( 91.66, 30.92);

\path[draw=drawColor,line width= 0.6pt,line join=round] (112.96, 28.17) --
	(112.96, 30.92);

\path[draw=drawColor,line width= 0.6pt,line join=round] (134.25, 28.17) --
	(134.25, 30.92);
\end{scope}
\begin{scope}
\path[clip] (  0.00,  0.00) rectangle (144.54,133.70);
\definecolor{drawColor}{RGB}{190,190,190}

\node[text=drawColor,anchor=base,inner sep=0pt, outer sep=0pt, scale=  0.79] at ( 38.43, 20.52) {.08};

\node[text=drawColor,anchor=base,inner sep=0pt, outer sep=0pt, scale=  0.79] at ( 49.08, 20.52) {.1};

\node[text=drawColor,anchor=base,inner sep=0pt, outer sep=0pt, scale=  0.79] at ( 70.37, 20.52) {.3};

\node[text=drawColor,anchor=base,inner sep=0pt, outer sep=0pt, scale=  0.79] at ( 91.66, 20.52) {.5};

\node[text=drawColor,anchor=base,inner sep=0pt, outer sep=0pt, scale=  0.79] at (112.96, 20.52) {.7};

\node[text=drawColor,anchor=base,inner sep=0pt, outer sep=0pt, scale=  0.79] at (134.25, 20.52) {9};
\end{scope}
\begin{scope}
\path[clip] (  0.00,  0.00) rectangle (144.54,133.70);
\definecolor{drawColor}{RGB}{0,0,0}

\node[text=drawColor,anchor=base,inner sep=0pt, outer sep=0pt, scale=  1.10] at ( 86.34,  7.44) {Sampling Rate};
\end{scope}
\begin{scope}
\path[clip] (  0.00,  0.00) rectangle (144.54,133.70);

\path[] ( 50.96, 74.17) rectangle (142.80,125.65);
\end{scope}
\begin{scope}
\path[clip] (  0.00,  0.00) rectangle (144.54,133.70);

\path[] ( 56.65, 94.31) rectangle ( 71.10,108.77);
\end{scope}
\begin{scope}
\path[clip] (  0.00,  0.00) rectangle (144.54,133.70);
\definecolor{drawColor}{RGB}{0,0,255}

\path[draw=drawColor,line width= 1.7pt,line join=round] ( 58.10,101.54) -- ( 69.66,101.54);
\end{scope}
\begin{scope}
\path[clip] (  0.00,  0.00) rectangle (144.54,133.70);
\definecolor{drawColor}{RGB}{0,0,255}

\path[draw=drawColor,line width= 1.7pt,line join=round] ( 58.10,101.54) -- ( 69.66,101.54);
\end{scope}
\begin{scope}
\path[clip] (  0.00,  0.00) rectangle (144.54,133.70);

\path[] ( 56.65, 79.86) rectangle ( 71.10, 94.31);
\end{scope}
\begin{scope}
\path[clip] (  0.00,  0.00) rectangle (144.54,133.70);
\definecolor{drawColor}{RGB}{0,0,0}

\path[draw=drawColor,line width= 1.7pt,line join=round] ( 58.10, 87.08) -- ( 69.66, 87.08);
\end{scope}
\begin{scope}
\path[clip] (  0.00,  0.00) rectangle (144.54,133.70);
\definecolor{drawColor}{RGB}{0,0,0}

\path[draw=drawColor,line width= 1.7pt,line join=round] ( 58.10, 87.08) -- ( 69.66, 87.08);
\end{scope}
\begin{scope}
\path[clip] (  0.00,  0.00) rectangle (144.54,133.70);
\definecolor{drawColor}{RGB}{0,0,0}

\node[text=drawColor,anchor=base west,inner sep=0pt, outer sep=0pt, scale=  0.88] at ( 72.91, 98.51) {{\sc Fsc} (this paper)};
\end{scope}
\begin{scope}
\path[clip] (  0.00,  0.00) rectangle (144.54,133.70);
\definecolor{drawColor}{RGB}{0,0,0}

\node[text=drawColor,anchor=base west,inner sep=0pt, outer sep=0pt, scale=  0.88] at ( 72.91, 84.05) {Best$(a$-$h)$};
\end{scope}
\begin{scope}
\path[clip] (  0.00,  0.00) rectangle (144.54,133.70);
\definecolor{drawColor}{RGB}{0,0,0}

\node[text=drawColor,anchor=base,inner sep=0pt, outer sep=0pt, scale=  1.10] at ( 86.34,120.62) {Hopkins - Missing Data};
\end{scope}
\end{tikzpicture}} \hspace{-.3cm}
\Scale[.6]{
\begin{tikzpicture}[x=1pt,y=1pt]
\definecolor{fillColor}{RGB}{255,255,255}
\path[use as bounding box,fill=fillColor,fill opacity=0.00] (0,0) rectangle (144.54,133.70);
\begin{scope}
\path[clip] (  0.00,  0.00) rectangle (144.54,133.70);
\definecolor{drawColor}{RGB}{255,255,255}
\definecolor{fillColor}{RGB}{255,255,255}

\path[draw=drawColor,line width= 0.6pt,line join=round,line cap=round,fill=fillColor] ( -0.00,  0.00) rectangle (144.54,133.70);
\end{scope}
\begin{scope}
\path[clip] ( 40.24, 30.92) rectangle (139.04,113.05);
\definecolor{fillColor}{gray}{0.92}

\path[fill=fillColor] ( 40.24, 30.92) rectangle (139.04,113.05);
\definecolor{drawColor}{RGB}{255,255,255}

\path[draw=drawColor,line width= 0.3pt,line join=round] ( 40.24, 47.10) --
	(139.04, 47.10);

\path[draw=drawColor,line width= 0.3pt,line join=round] ( 40.24, 71.99) --
	(139.04, 71.99);

\path[draw=drawColor,line width= 0.3pt,line join=round] ( 40.24, 96.88) --
	(139.04, 96.88);

\path[draw=drawColor,line width= 0.3pt,line join=round] ( 55.96, 30.92) --
	( 55.96,113.05);

\path[draw=drawColor,line width= 0.3pt,line join=round] ( 78.41, 30.92) --
	( 78.41,113.05);

\path[draw=drawColor,line width= 0.3pt,line join=round] (100.87, 30.92) --
	(100.87,113.05);

\path[draw=drawColor,line width= 0.3pt,line join=round] (123.32, 30.92) --
	(123.32,113.05);

\path[draw=drawColor,line width= 0.6pt,line join=round] ( 40.24, 34.66) --
	(139.04, 34.66);

\path[draw=drawColor,line width= 0.6pt,line join=round] ( 40.24, 59.54) --
	(139.04, 59.54);

\path[draw=drawColor,line width= 0.6pt,line join=round] ( 40.24, 84.43) --
	(139.04, 84.43);

\path[draw=drawColor,line width= 0.6pt,line join=round] ( 40.24,109.32) --
	(139.04,109.32);

\path[draw=drawColor,line width= 0.6pt,line join=round] ( 44.73, 30.92) --
	( 44.73,113.05);

\path[draw=drawColor,line width= 0.6pt,line join=round] ( 67.19, 30.92) --
	( 67.19,113.05);

\path[draw=drawColor,line width= 0.6pt,line join=round] ( 89.64, 30.92) --
	( 89.64,113.05);

\path[draw=drawColor,line width= 0.6pt,line join=round] (112.10, 30.92) --
	(112.10,113.05);

\path[draw=drawColor,line width= 0.6pt,line join=round] (134.55, 30.92) --
	(134.55,113.05);
\definecolor{drawColor}{RGB}{0,0,0}

\path[draw=drawColor,line width= 1.7pt,line join=round] ( 44.73, 59.54) --
	( 67.19, 61.10) --
	( 89.64, 60.71) --
	(112.10, 60.48) --
	(134.55, 59.54);
\definecolor{drawColor}{RGB}{0,0,255}

\path[draw=drawColor,line width= 1.7pt,line join=round] ( 44.73, 62.66) --
	( 67.19, 65.87) --
	( 89.64, 63.53) --
	(112.10, 64.22) --
	(134.55, 65.72);
\end{scope}
\begin{scope}
\path[clip] (  0.00,  0.00) rectangle (144.54,133.70);
\definecolor{drawColor}{RGB}{190,190,190}

\node[text=drawColor,anchor=base,inner sep=0pt, outer sep=0pt, scale=  0.79] at ( 26.93, 31.93) {-0.25};

\node[text=drawColor,anchor=base,inner sep=0pt, outer sep=0pt, scale=  0.79] at ( 26.93, 56.82) {0.00};

\node[text=drawColor,anchor=base,inner sep=0pt, outer sep=0pt, scale=  0.79] at ( 26.93, 81.70) {0.25};

\node[text=drawColor,anchor=base,inner sep=0pt, outer sep=0pt, scale=  0.79] at ( 26.93,106.59) {0.50};
\end{scope}
\begin{scope}
\path[clip] (  0.00,  0.00) rectangle (144.54,133.70);
\definecolor{drawColor}{gray}{0.20}

\path[draw=drawColor,line width= 0.6pt,line join=round] ( 37.49, 34.66) --
	( 40.24, 34.66);

\path[draw=drawColor,line width= 0.6pt,line join=round] ( 37.49, 59.54) --
	( 40.24, 59.54);

\path[draw=drawColor,line width= 0.6pt,line join=round] ( 37.49, 84.43) --
	( 40.24, 84.43);

\path[draw=drawColor,line width= 0.6pt,line join=round] ( 37.49,109.32) --
	( 40.24,109.32);
\end{scope}
\begin{scope}
\path[clip] (  0.00,  0.00) rectangle (144.54,133.70);
\definecolor{drawColor}{gray}{0.20}

\path[draw=drawColor,line width= 0.6pt,line join=round] ( 44.73, 28.17) --
	( 44.73, 30.92);

\path[draw=drawColor,line width= 0.6pt,line join=round] ( 67.19, 28.17) --
	( 67.19, 30.92);

\path[draw=drawColor,line width= 0.6pt,line join=round] ( 89.64, 28.17) --
	( 89.64, 30.92);

\path[draw=drawColor,line width= 0.6pt,line join=round] (112.10, 28.17) --
	(112.10, 30.92);

\path[draw=drawColor,line width= 0.6pt,line join=round] (134.55, 28.17) --
	(134.55, 30.92);
\end{scope}
\begin{scope}
\path[clip] (  0.00,  0.00) rectangle (144.54,133.70);
\definecolor{drawColor}{RGB}{190,190,190}

\node[text=drawColor,anchor=base,inner sep=0pt, outer sep=0pt, scale=  0.79] at ( 44.73, 20.52) {2};

\node[text=drawColor,anchor=base,inner sep=0pt, outer sep=0pt, scale=  0.79] at ( 67.19, 20.52) {3};

\node[text=drawColor,anchor=base,inner sep=0pt, outer sep=0pt, scale=  0.79] at ( 89.64, 20.52) {4};

\node[text=drawColor,anchor=base,inner sep=0pt, outer sep=0pt, scale=  0.79] at (112.10, 20.52) {5};

\node[text=drawColor,anchor=base,inner sep=0pt, outer sep=0pt, scale=  0.79] at (134.55, 20.52) {6};
\end{scope}
\begin{scope}
\path[clip] (  0.00,  0.00) rectangle (144.54,133.70);
\definecolor{drawColor}{RGB}{0,0,0}

\node[text=drawColor,anchor=base,inner sep=0pt, outer sep=0pt, scale=  1.10] at ( 89.64,  7.44) {Number of People};
\end{scope}
\begin{scope}
\path[clip] (  0.00,  0.00) rectangle (144.54,133.70);
\definecolor{drawColor}{RGB}{0,0,0}

\node[text=drawColor,anchor=base,inner sep=0pt, outer sep=0pt, scale=  1.10] at ( 89.64,120.62) {Yale - Full Data};
\end{scope}
\end{tikzpicture}} \hspace{-.3cm}
\Scale[.6]{
\begin{tikzpicture}[x=1pt,y=1pt]
\definecolor{fillColor}{RGB}{255,255,255}
\path[use as bounding box,fill=fillColor,fill opacity=0.00] (0,0) rectangle (144.54,133.70);
\begin{scope}
\path[clip] (  0.00,  0.00) rectangle (144.54,133.70);
\definecolor{drawColor}{RGB}{255,255,255}
\definecolor{fillColor}{RGB}{255,255,255}

\path[draw=drawColor,line width= 0.6pt,line join=round,line cap=round,fill=fillColor] (  0.00,  0.00) rectangle (144.54,133.70);
\end{scope}
\begin{scope}
\path[clip] ( 33.64, 30.92) rectangle (139.04,112.08);
\definecolor{fillColor}{gray}{0.92}

\path[fill=fillColor] ( 33.64, 30.92) rectangle (139.04,112.08);
\definecolor{drawColor}{RGB}{255,255,255}

\path[draw=drawColor,line width= 0.3pt,line join=round] ( 33.64, 45.08) --
	(139.04, 45.08);

\path[draw=drawColor,line width= 0.3pt,line join=round] ( 33.64, 66.01) --
	(139.04, 66.01);

\path[draw=drawColor,line width= 0.3pt,line join=round] ( 33.64, 86.94) --
	(139.04, 86.94);

\path[draw=drawColor,line width= 0.3pt,line join=round] ( 33.64,107.87) --
	(139.04,107.87);

\path[draw=drawColor,line width= 0.3pt,line join=round] ( 43.76, 30.92) --
	( 43.76,112.08);

\path[draw=drawColor,line width= 0.3pt,line join=round] ( 59.73, 30.92) --
	( 59.73,112.08);

\path[draw=drawColor,line width= 0.3pt,line join=round] ( 81.02, 30.92) --
	( 81.02,112.08);

\path[draw=drawColor,line width= 0.3pt,line join=round] (102.31, 30.92) --
	(102.31,112.08);

\path[draw=drawColor,line width= 0.3pt,line join=round] (123.60, 30.92) --
	(123.60,112.08);

\path[draw=drawColor,line width= 0.6pt,line join=round] ( 33.64, 34.61) --
	(139.04, 34.61);

\path[draw=drawColor,line width= 0.6pt,line join=round] ( 33.64, 55.54) --
	(139.04, 55.54);

\path[draw=drawColor,line width= 0.6pt,line join=round] ( 33.64, 76.47) --
	(139.04, 76.47);

\path[draw=drawColor,line width= 0.6pt,line join=round] ( 33.64, 97.40) --
	(139.04, 97.40);

\path[draw=drawColor,line width= 0.6pt,line join=round] ( 38.43, 30.92) --
	( 38.43,112.08);

\path[draw=drawColor,line width= 0.6pt,line join=round] ( 49.08, 30.92) --
	( 49.08,112.08);

\path[draw=drawColor,line width= 0.6pt,line join=round] ( 70.37, 30.92) --
	( 70.37,112.08);

\path[draw=drawColor,line width= 0.6pt,line join=round] ( 91.66, 30.92) --
	( 91.66,112.08);

\path[draw=drawColor,line width= 0.6pt,line join=round] (112.96, 30.92) --
	(112.96,112.08);

\path[draw=drawColor,line width= 0.6pt,line join=round] (134.25, 30.92) --
	(134.25,112.08);
\definecolor{drawColor}{RGB}{0,0,0}

\path[draw=drawColor,line width= 1.7pt,line join=round] ( 38.43,108.39) --
	( 49.08,107.65) --
	( 59.73, 92.39) --
	( 70.37, 79.31) --
	( 81.02, 55.32) --
	( 91.66, 46.61) --
	(102.31, 38.98) --
	(112.96, 34.61) --
	(123.60, 34.61) --
	(134.25, 34.61);
\definecolor{drawColor}{RGB}{0,0,255}

\path[draw=drawColor,line width= 1.7pt,line join=round] ( 38.43,105.77) --
	( 49.08, 61.51) --
	( 59.73, 58.68) --
	( 70.37, 56.72) --
	( 81.02, 54.51) --
	( 91.66, 48.74) --
	(102.31, 41.23) --
	(112.96, 39.85) --
	(123.60, 39.32) --
	(134.25, 38.54);
\end{scope}
\begin{scope}
\path[clip] (  0.00,  0.00) rectangle (144.54,133.70);
\definecolor{drawColor}{RGB}{190,190,190}

\node[text=drawColor,anchor=base,inner sep=0pt, outer sep=0pt, scale=  0.79] at ( 23.63, 31.89) {0.0};

\node[text=drawColor,anchor=base,inner sep=0pt, outer sep=0pt, scale=  0.79] at ( 23.63, 52.82) {0.2};

\node[text=drawColor,anchor=base,inner sep=0pt, outer sep=0pt, scale=  0.79] at ( 23.63, 73.75) {0.4};

\node[text=drawColor,anchor=base,inner sep=0pt, outer sep=0pt, scale=  0.79] at ( 23.63, 94.68) {0.6};
\end{scope}
\begin{scope}
\path[clip] (  0.00,  0.00) rectangle (144.54,133.70);
\definecolor{drawColor}{gray}{0.20}

\path[draw=drawColor,line width= 0.6pt,line join=round] ( 30.89, 34.61) --
	( 33.64, 34.61);

\path[draw=drawColor,line width= 0.6pt,line join=round] ( 30.89, 55.54) --
	( 33.64, 55.54);

\path[draw=drawColor,line width= 0.6pt,line join=round] ( 30.89, 76.47) --
	( 33.64, 76.47);

\path[draw=drawColor,line width= 0.6pt,line join=round] ( 30.89, 97.40) --
	( 33.64, 97.40);
\end{scope}
\begin{scope}
\path[clip] (  0.00,  0.00) rectangle (144.54,133.70);
\definecolor{drawColor}{gray}{0.20}

\path[draw=drawColor,line width= 0.6pt,line join=round] ( 38.43, 28.17) --
	( 38.43, 30.92);

\path[draw=drawColor,line width= 0.6pt,line join=round] ( 49.08, 28.17) --
	( 49.08, 30.92);

\path[draw=drawColor,line width= 0.6pt,line join=round] ( 70.37, 28.17) --
	( 70.37, 30.92);

\path[draw=drawColor,line width= 0.6pt,line join=round] ( 91.66, 28.17) --
	( 91.66, 30.92);

\path[draw=drawColor,line width= 0.6pt,line join=round] (112.96, 28.17) --
	(112.96, 30.92);

\path[draw=drawColor,line width= 0.6pt,line join=round] (134.25, 28.17) --
	(134.25, 30.92);
\end{scope}
\begin{scope}
\path[clip] (  0.00,  0.00) rectangle (144.54,133.70);
\definecolor{drawColor}{RGB}{190,190,190}

\node[text=drawColor,anchor=base,inner sep=0pt, outer sep=0pt, scale=  0.79] at ( 38.43, 20.52) {$10^{-2}$};

\node[text=drawColor,anchor=base,inner sep=0pt, outer sep=0pt, scale=  0.79] at ( 49.08, 20.52) {.1};

\node[text=drawColor,anchor=base,inner sep=0pt, outer sep=0pt, scale=  0.79] at ( 70.37, 20.52) {.3};

\node[text=drawColor,anchor=base,inner sep=0pt, outer sep=0pt, scale=  0.79] at ( 91.66, 20.52) {.5};

\node[text=drawColor,anchor=base,inner sep=0pt, outer sep=0pt, scale=  0.79] at (112.96, 20.52) {.7};

\node[text=drawColor,anchor=base,inner sep=0pt, outer sep=0pt, scale=  0.79] at (134.25, 20.52) {.9};
\end{scope}
\begin{scope}
\path[clip] (  0.00,  0.00) rectangle (144.54,133.70);
\definecolor{drawColor}{RGB}{0,0,0}

\node[text=drawColor,anchor=base,inner sep=0pt, outer sep=0pt, scale=  1.10] at ( 86.34,  7.44) {Sampling Rate};
\end{scope}
\begin{scope}
\path[clip] (  0.00,  0.00) rectangle (144.54,133.70);
\definecolor{drawColor}{RGB}{0,0,0}

\node[text=drawColor,anchor=base,inner sep=0pt, outer sep=0pt, scale=  1.10] at ( 86.34,120.62) {Yale - Missing Data};
\end{scope}
\end{tikzpicture}}
\caption{Clustering error on real datasets. Comparing \FSC\ vs.~the best algorithm among $(a)$-$(h)$ in each trial. Notice the different scales. With full-data \FSC\ is rarely outperformed by other algorithms, and only by a marginal amount (only about $3\%$ lower than the best among $(a)$-$(h)$). In contrast, when data is missing, \FSC\ outperforms other algorithms by a wide margin. For example, in the Yale B experiment with $\p=0.1$, \FSC\ gets $25.7\%$ error, while the next best algorithm gets $69.79\%$.}
\label{realDataFig}
\end{figure}

First we test \FSC\ with full data. Figure \ref{realDataFig} shows the results of 10 videos. We can see a consistent behavior between \FSC\ and the state-of-the-art. Pay attention to the scale, showing that \FSC's accuracy is only about $3\%$ lower than the best among $(a)$-$(h)$. Figure \ref{realDataFig} also shows the average clustering error (of all videos) as a function of the amount of missing data (induced uniformly at random). Consistent with our simulations, \FSC\ dramatically outperforms the state-of-the-art in the low-sampling regime (many missing data); for example, with $\p=0.1$, the best among algorithms $(a)$-$(h)$ gets $52.95\%$ error, which is close to random guessing (because on average, there are $\K=2$ subspaces in each video). In contrast, \FSC\ achieves $15.03\%$ error. Notice that $\p=0.1$ is very close to the exact information-theoretic minimum sampling rate $\p=(\r+1)/\d=0.086$ \cite{infoTheoretic}.

\begin{figure}
\centering
\begin{tabular}{cccccc}
100\% \hspace{.8cm}
& 90\% \hspace{.8cm}
& 80\% \hspace{.8cm}
& 60\% \hspace{.8cm}
& 40\% \hspace{.8cm}
& 20\%
\end{tabular} \\
\includegraphics[height=1.9cm]{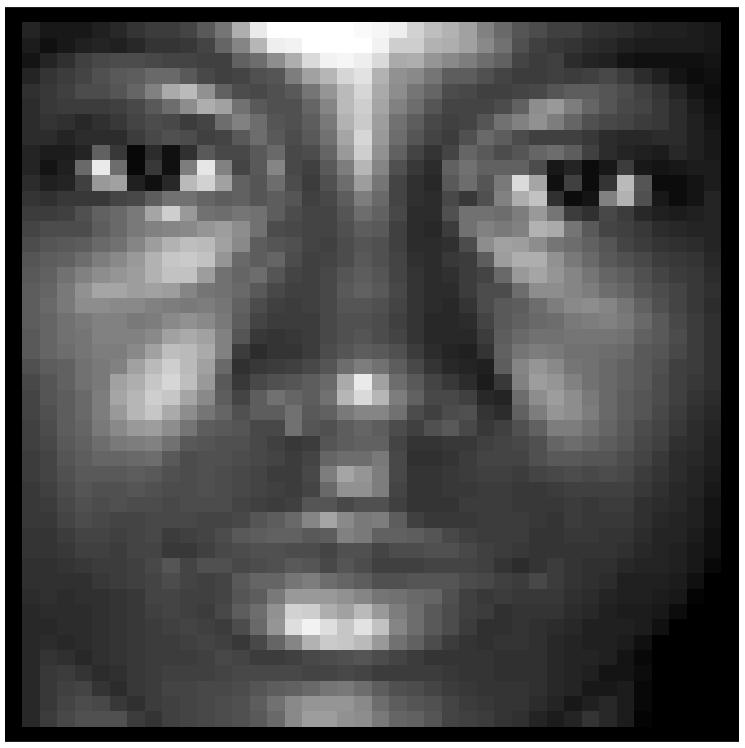}
\includegraphics[height=1.9cm]{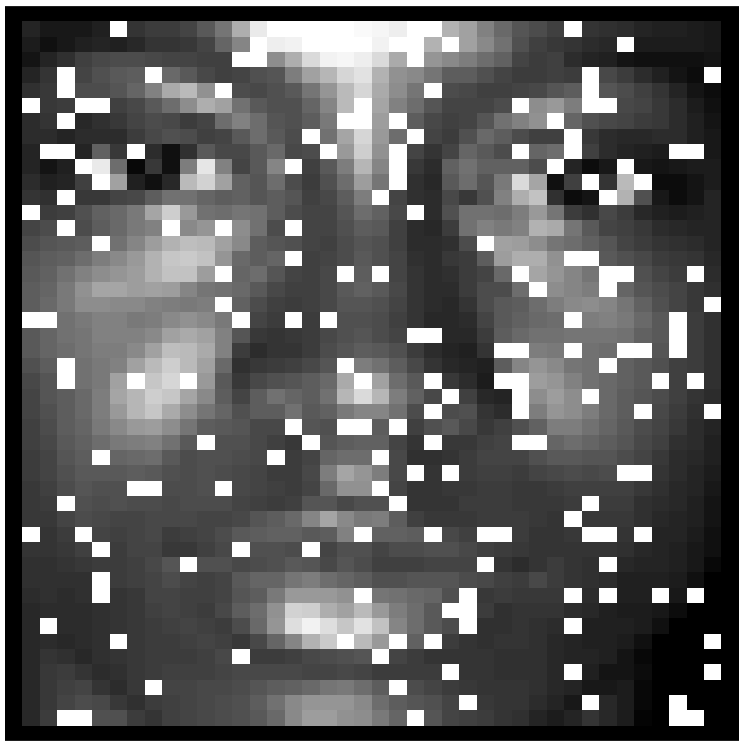}
\includegraphics[height=1.9cm]{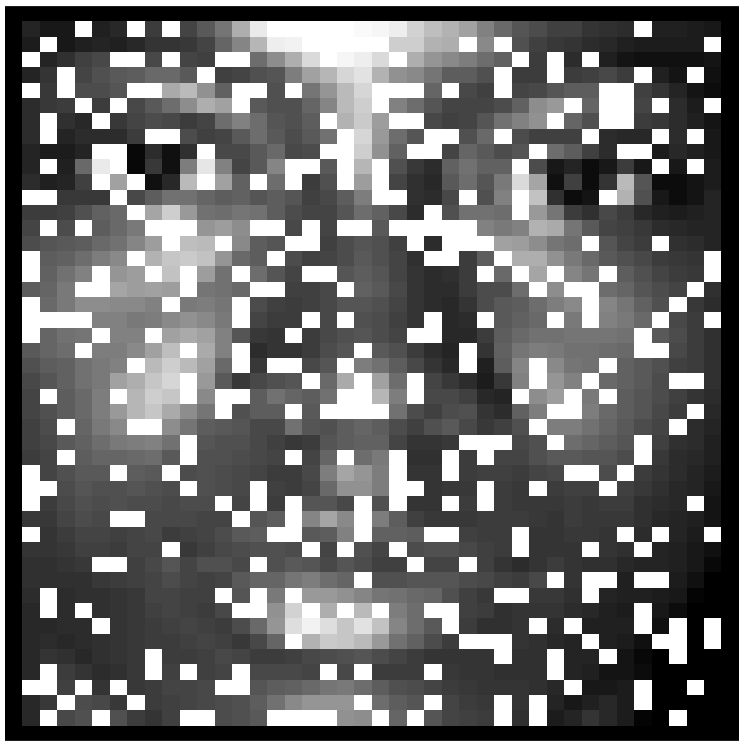}
\includegraphics[height=1.9cm]{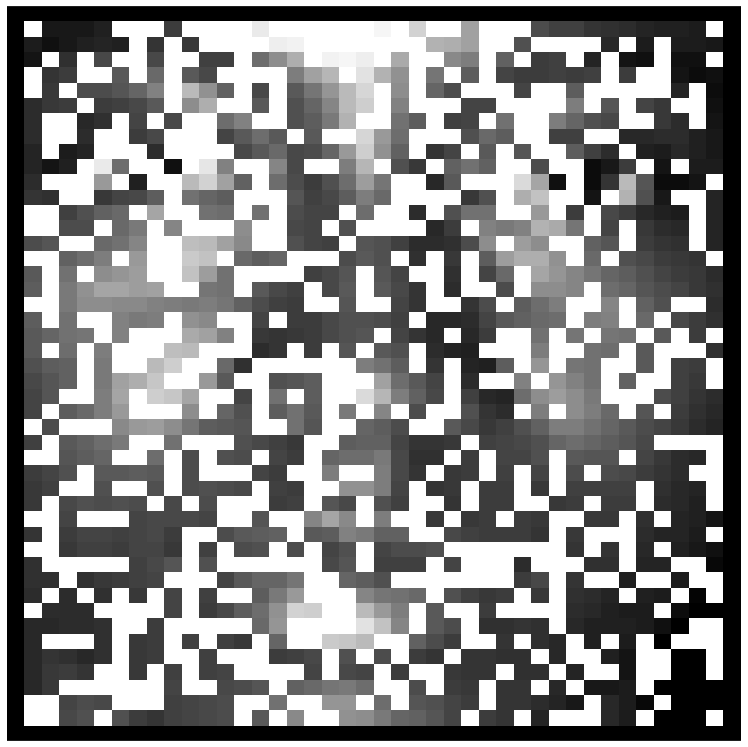}
\includegraphics[height=1.9cm]{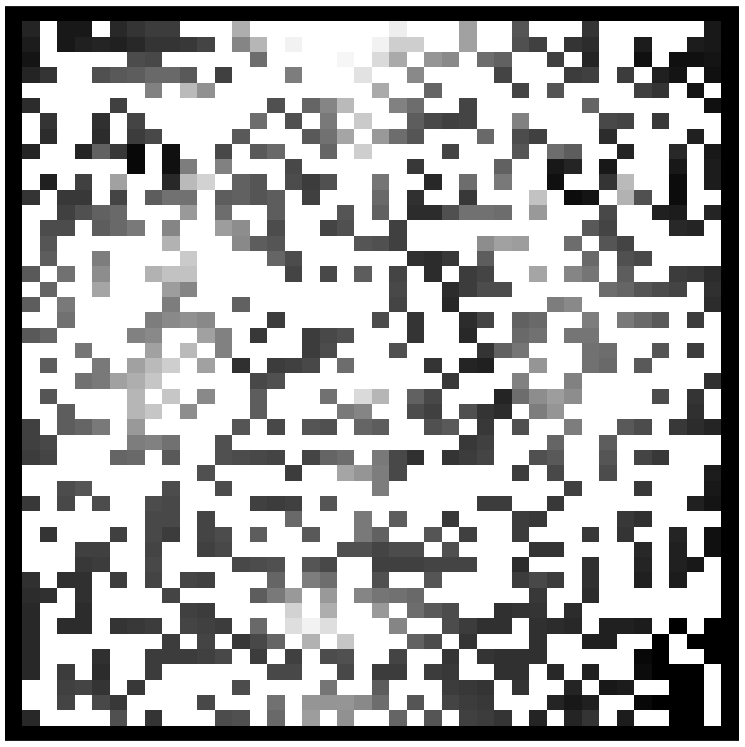}
\includegraphics[height=1.9cm]{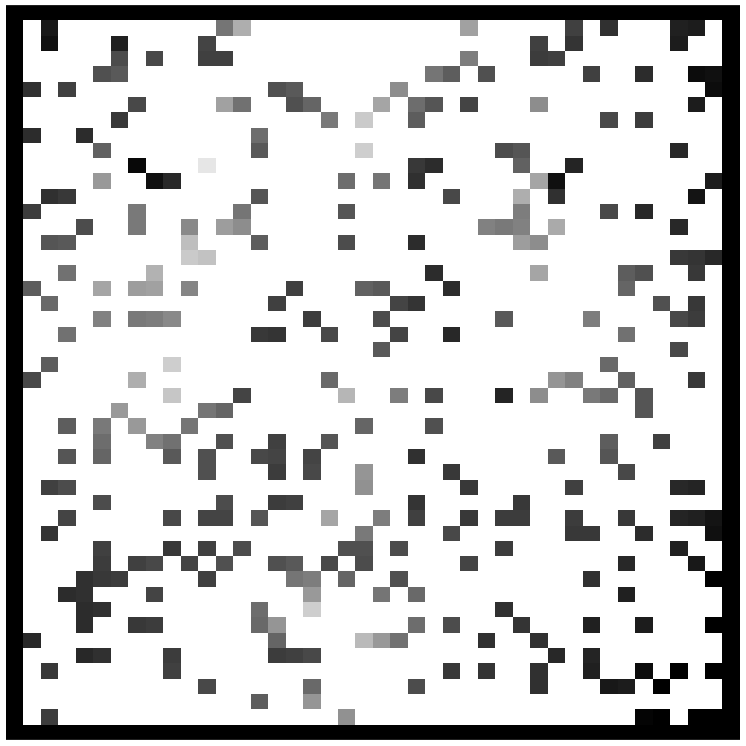}
\\
\includegraphics[height=1.9cm]{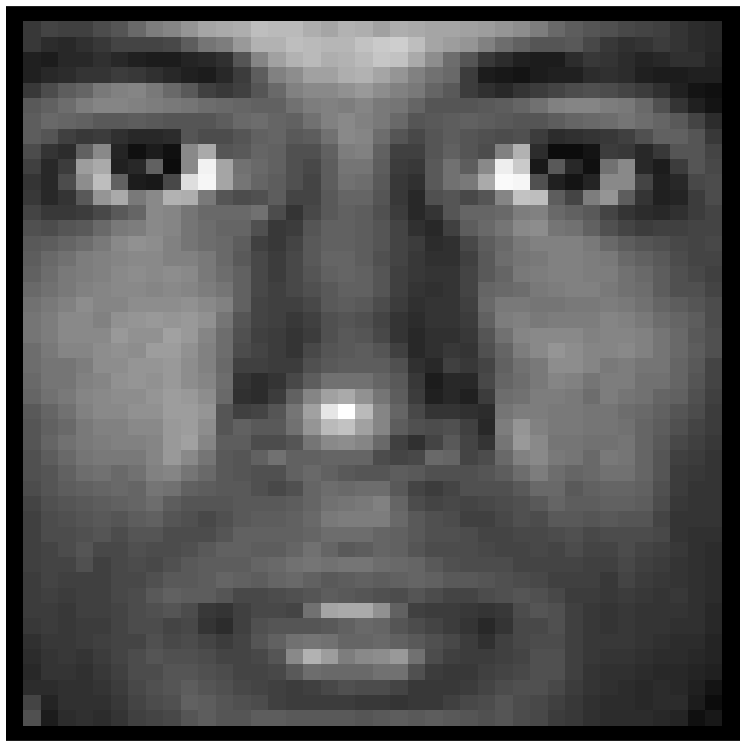}
\includegraphics[height=1.9cm]{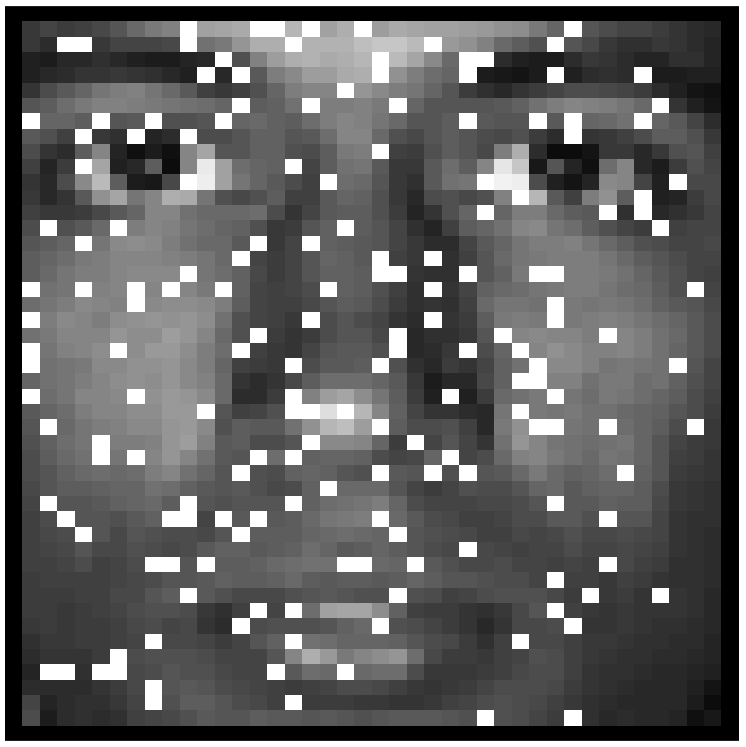}
\includegraphics[height=1.9cm]{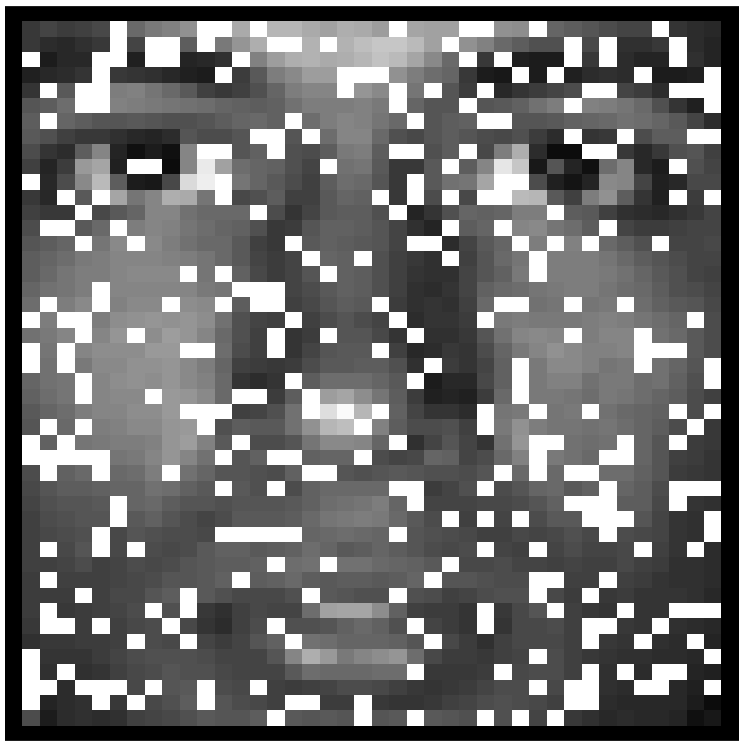}
\includegraphics[height=1.9cm]{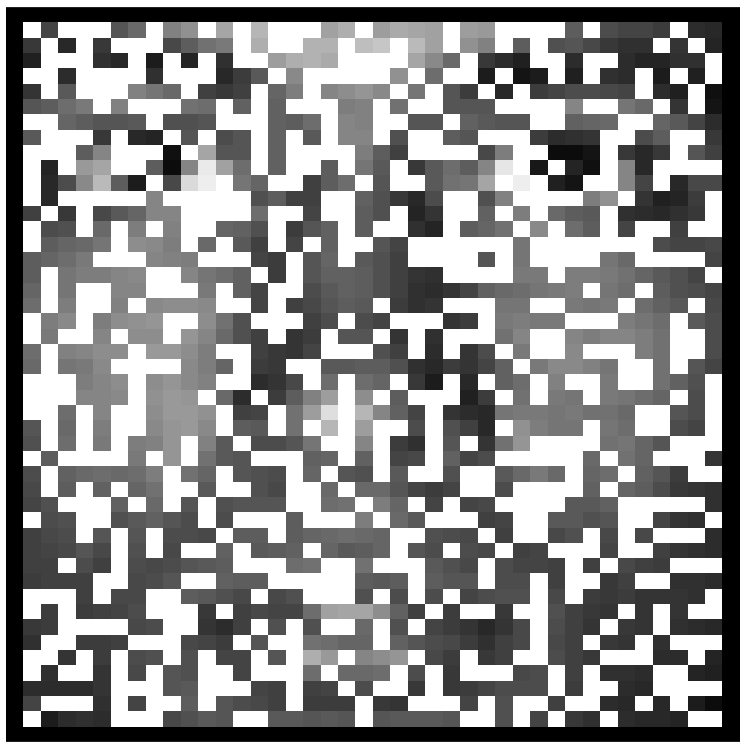}
\includegraphics[height=1.9cm]{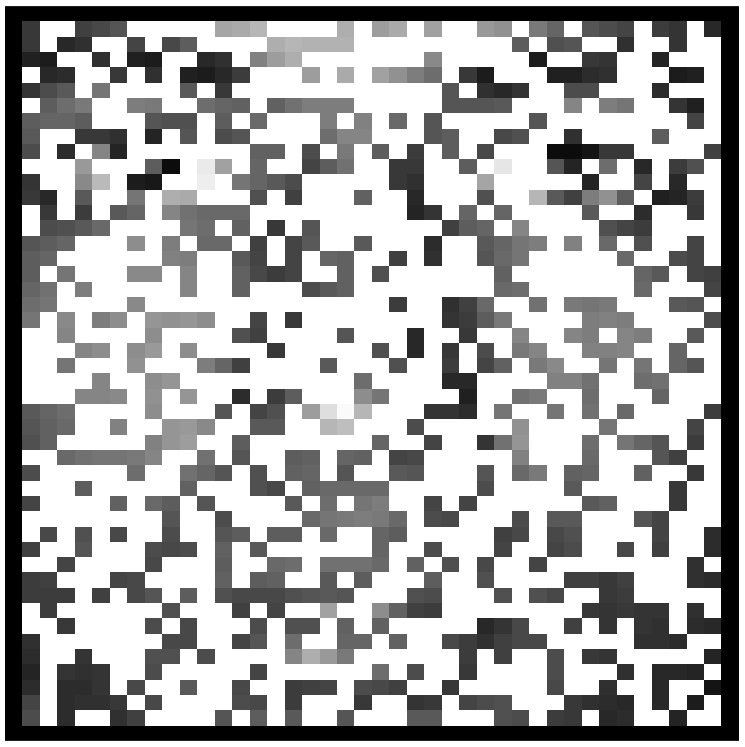}
\includegraphics[height=1.9cm]{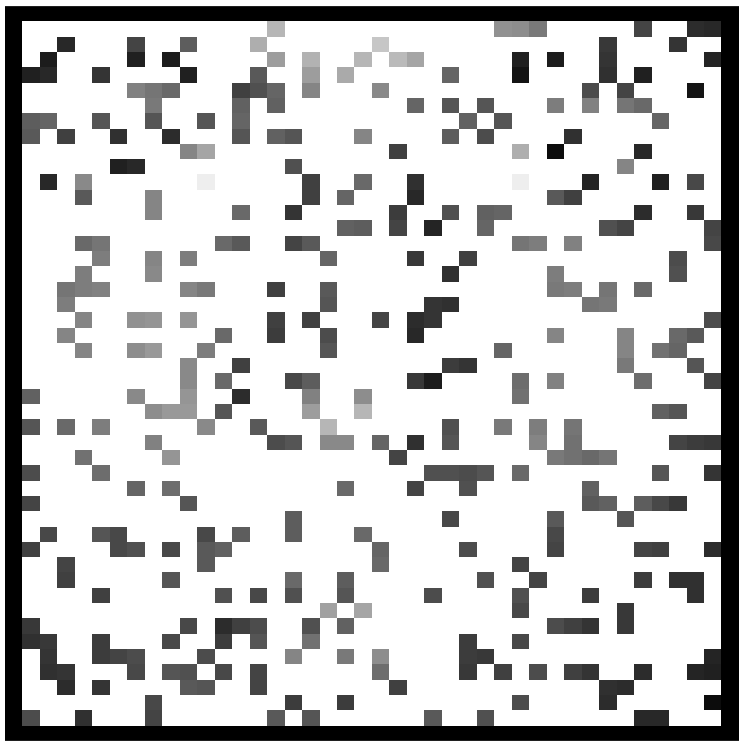}
\\
\includegraphics[height=1.9cm]{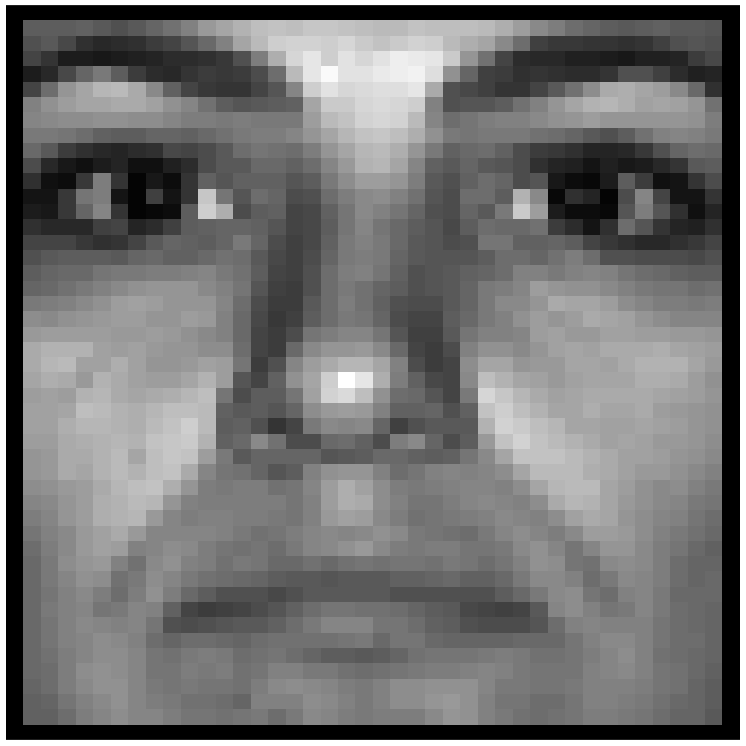}
\includegraphics[height=1.9cm]{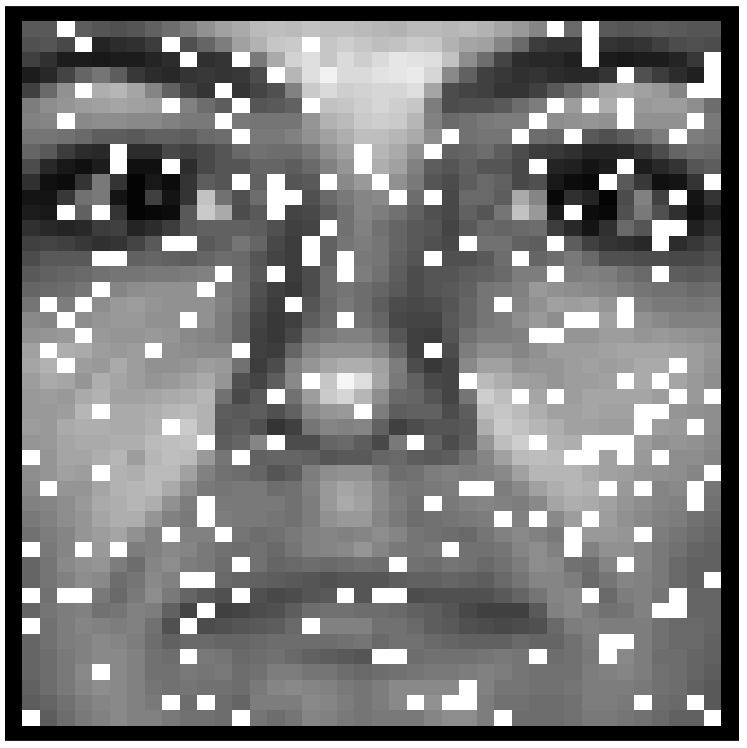}
\includegraphics[height=1.9cm]{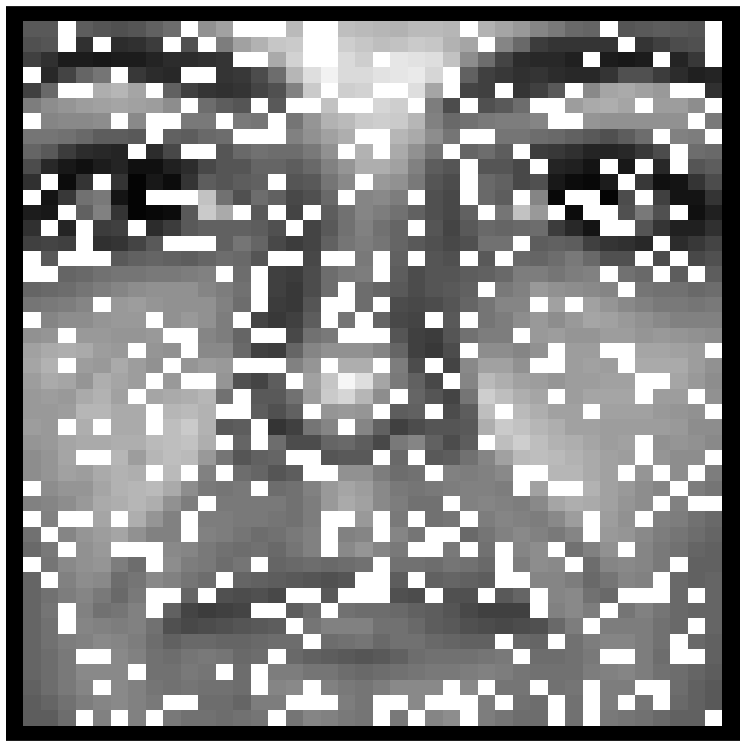}
\includegraphics[height=1.9cm]{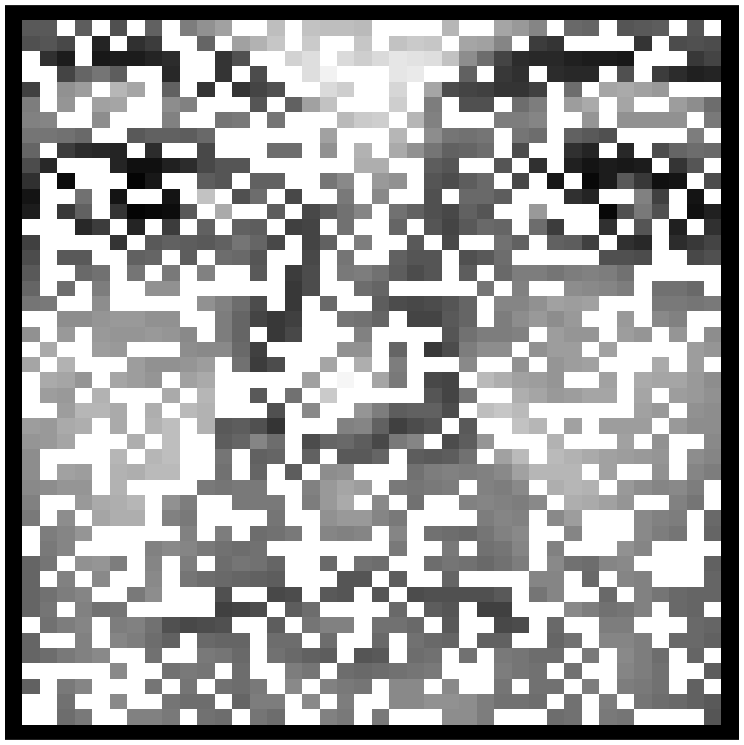}
\includegraphics[height=1.9cm]{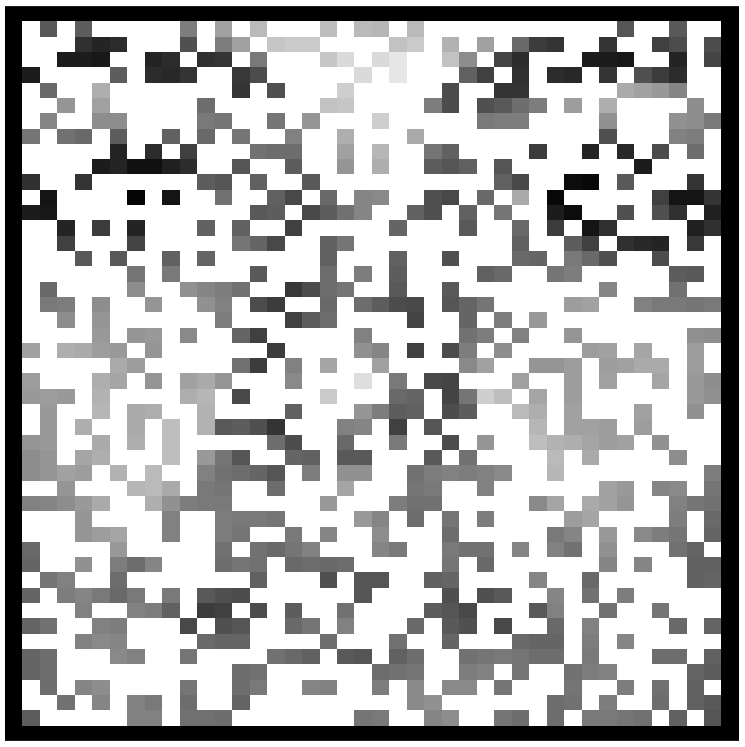}
\includegraphics[height=1.9cm]{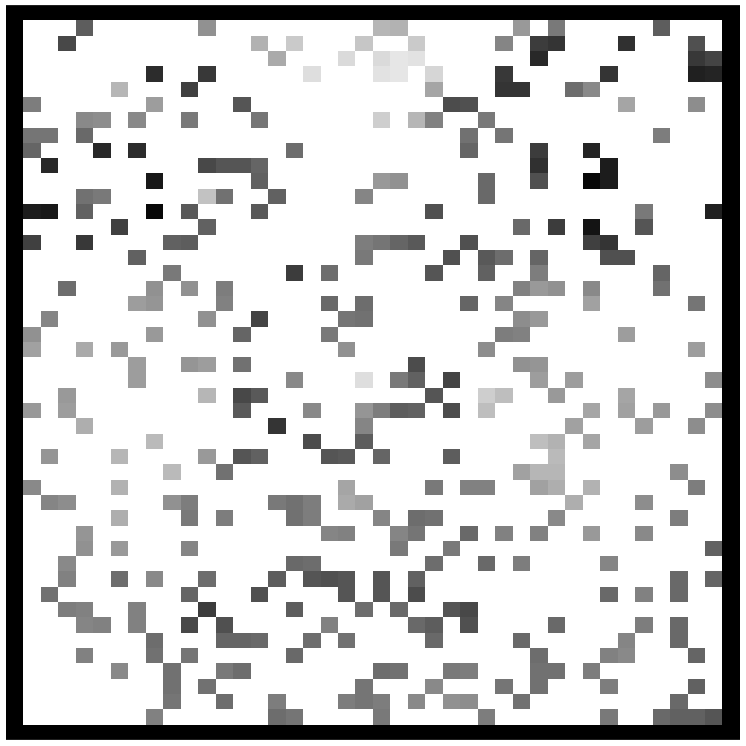}
\caption{Sample images from the Yale B dataset \cite{yale} under decreasing sampling rates $\p$. \FSC\ achieves $\approx75\%$ clustering accuracy with as little as $\p=10\%$ observations. The next best algorithm only achieves $\approx30\%$.}
\label{yaleFig}
\end{figure}

{\bf Face Clustering.} It has been shown that the vectorized images of the same person under different illuminations lie near a $9$-dimensonal subspace \cite{lambertian}. In this experiment we evaluate the performance of \FSC\ at clustering faces of multiple individuals, using the Yale B dataset \cite{yale}, containing a total of $2432$ images, each of size $48 \times 42$, evenly distributed amongst $38$ individuals; Figure \ref{yaleFig} contains a few samples. To compare things vis \`a vis, before clustering, we use robust \PCA\ \cite{robustpca} on each cluster, to remove outliers; this is a widely used preprocessing step \cite{ssc,ewzf,ssp14,elhamifar}. In each of $30$ trials, we select $\K$ people uniformly at random, and record the clustering error. Figure \ref{realDataFig} shows that \FSC\ is quite competitive. Notice that there is only a small gap of $5\%$ error between \FSC\ and the best algorithm (for each trial) amongst $(a)$-$(h)$, which in most cases was \SSC. Figure \ref{realDataFig} also shows the average clustering error as a function of the amount of missing data (induced uniformly at random), with $\K$ fixed to $6$ people. Again, \FSC\ outperforms the state-of-the-art in the low-sampling regime (many missing data). For example, with $\p=0.1$ \FSC\ gets $25.7\%$ error, while the next best algorithm gets $69.79\%$. Notice that $\p=0.1$ is quite close to the exact information-theoretic necessary $\p=(\r+1)/\d=0.005$ \cite{infoTheoretic}.

\section{Future Directions}
This paper introduces \FSC, and shows its great potential to handle missing data. However, the term $\U{}_\i^\T\U_\i$ in $\P_\i$ may become ill-conditioned (especially if the rank is over-estimated), and its inversion is computationally expensive. Also notice that the storage and computational complexity of \FSC\ is polynomial in the number of data points $\n$. Exciting directions for future work that are out of the scope of this paper will address these caveats using updates over the Grassmannian, as is done in \cite{grouse} for subspace tracking, as well as greedy, adaptive, and data-driven variants that quickly fuse promising subspaces within a confidence interval, in order to improve complexity.

\newpage
\bibliographystyle{nihunsrt} 

\end{document}